\pdfoutput=1 
\documentclass{article}

\PassOptionsToPackage{numbers, compress}{natbib}
\usepackage[final]{neurips_2020}

\usepackage[utf8]{inputenc} 
\usepackage[T1]{fontenc}    
\usepackage{hyperref}       
\usepackage{url}            
\usepackage{booktabs}       
\usepackage{amsfonts}       
\usepackage{nicefrac}       
\usepackage{microtype}      
\usepackage{graphicx}
\usepackage{amsmath}
\usepackage{amsthm}
\usepackage{amssymb}
\usepackage{color}
\usepackage{overpic}
\usepackage{makecell}
\usepackage{float}
\usepackage{wrapfig}
\usepackage{tabularx}
\usepackage{siunitx}
\usepackage[noend]{algpseudocode}
\usepackage[ruled,vlined]{algorithm2e}
\SetKwBlock{Forward}{forward pass:}{end}
\SetKwBlock{Backward}{backward pass:}{end}
\makeatletter
\def\BState{\State\hskip-\ALG@thistlm}
\makeatother
\newtheorem{theorem}{Theorem}
\usepackage{multirow}

\title{MeshSDF: Differentiable Iso-Surface Extraction}

\newcommand*\samethanks[1][\value{footnote}]{\footnotemark[#1]}

\author{%
  Edoardo Remelli \thanks{Equal contribution} $^{~1}$ \\
  \And
  Artem Lukoianov \samethanks~$^{~1,2}$\\
  \And
  Stephan R. Richter $^{3}$ \\
  \AND
  Benoît Guillard $^{1}$\\
  \And
  Timur Bagautdinov $^{2}$\\
  \And
  Pierre Baque $^{2}$\\
  \And 
  Pascal Fua $^{1}$\\
  \AND
  $^{1}$\texttt{CVLab, EPFL,} \texttt{\{name.surname\}@epfl.ch} \\
  $^{2}$\texttt{Neural Concept SA,} \texttt{\{name.surname\}@neuralconcept.com} \\
  $^{3}$\texttt{Intel Labs,} \texttt{\{name.surname\}@intel.com}
}

\newif\ifdraft
\draftfalse
\drafttrue

\definecolor{orange}{rgb}{1,0.5,0}
\definecolor{violet}{RGB}{70,0,170}
\definecolor{pink}{RGB}{252,107,252}
\definecolor{brown}{RGB}{139,69,19}
\definecolor{red_fig}{RGB}{189,9,9}

 \newcommand{\ER}[1]{{\color{violet}{\bf ER: #1}}}
 
 \newcommand{\AL}[1]{{\color{blue}{\bf AL: #1}}}
 
 \newcommand{\SR}[1]{{\color{green}{\bf SR: #1}}}

\newcommand{\comment}[1]{}

\newcommand{\bq}{\mathbf{q}}
\newcommand{\bv}{\mathbf{v}}
\newcommand{\bz}{\mathbf{z}}
\newcommand{\bx}{\mathbf{x}}

\newcommand{\myfootnotesize}{\fontsize{8pt}{10pt}\selectfont}
\begin{document}

\maketitle

\begin{abstract}

Geometric Deep Learning has recently made striking progress with the advent of \textit{continuous} Deep Implicit Fields. They allow for detailed modeling of watertight surfaces of arbitrary topology while not relying on a 3D Euclidean grid, resulting in a learnable parameterization that is not limited in resolution. 

Unfortunately, these methods are often not suitable for applications that require an \textit{explicit} mesh-based surface representation because converting an implicit field to such a representation relies on the Marching Cubes algorithm, which cannot be differentiated with respect to the underlying implicit field.

In this work, we remove this limitation and introduce a differentiable way to produce explicit surface mesh representations from Deep Signed
Distance Functions. Our key insight is that by reasoning on how
implicit field perturbations impact local surface geometry, one
can ultimately differentiate the 3D location of surface samples 
with respect to the underlying deep implicit field.
We exploit this to define \textit{MeshSDF}, an end-to-end differentiable mesh representation which can vary its topology.

We use two different applications to validate our theoretical insight: Single-View Reconstruction via Differentiable Rendering and Physically-Driven Shape Optimization. In both cases our differentiable parameterization gives us an edge over state-of-the-art algorithms.
  
\end{abstract}

\section{Introduction}

Geometric Deep Learning has recently witnessed a breakthrough with the advent of \textit{continuous} Deep Implicit Fields \cite{Park20a,Mescheder19,Chen19c}. These enable detailed modeling of watertight surfaces, while not relying on a 3D Euclidean grid or meshes with fixed topology, resulting in a learnable surface parameterization that is \textit{not} limited in resolution.

However, a number of important applications require {\it explicit} surface representations, such as triangulated meshes or 3D point clouds. 
Computational Fluid Dynamics (CFD) simulations and the associated learning-based surrogate methods used for shape design in many engineering fields~\cite{Baque18,Umetani18} are a good example of this where 3D meshes serve as boundary conditions for the Navier-Stokes Equations. Similarly, many advanced physically-based rendering engines require surface meshes to model the complex interactions of light and physical surfaces efficiently~\cite{Nimier19,Pharr16}.

Combining explicit representations with the benefits of deep implicit fields requires converting the implicit surface parameterization to an explicit representation, which typically relies on one of the many variants of the Marching Cubes algorithm~\cite{Lorensen87,Newman06}. However, these approaches are not fully differentiable~\cite{Liao18a}. This effectively prevents the use of continuous Deep Implicit Fields as parameterizations when operating on explicit surface meshes. 

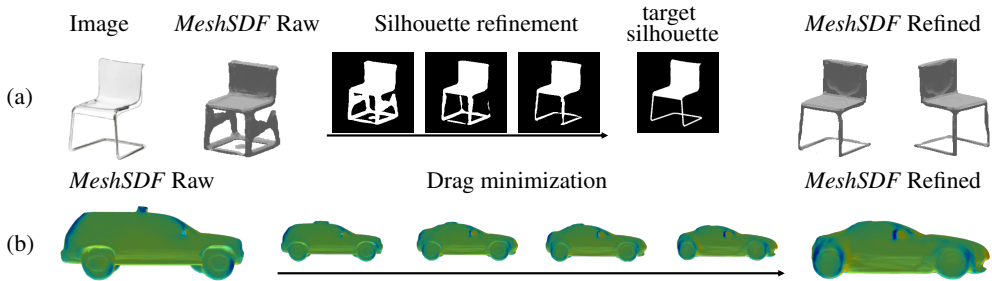
\begin{figure}[t]
        \vspace{-8pt}
		\begin{center}
			\begin{overpic}[clip, trim=0.0cm 10cm 0 0.5cm,width=1.0\textwidth]{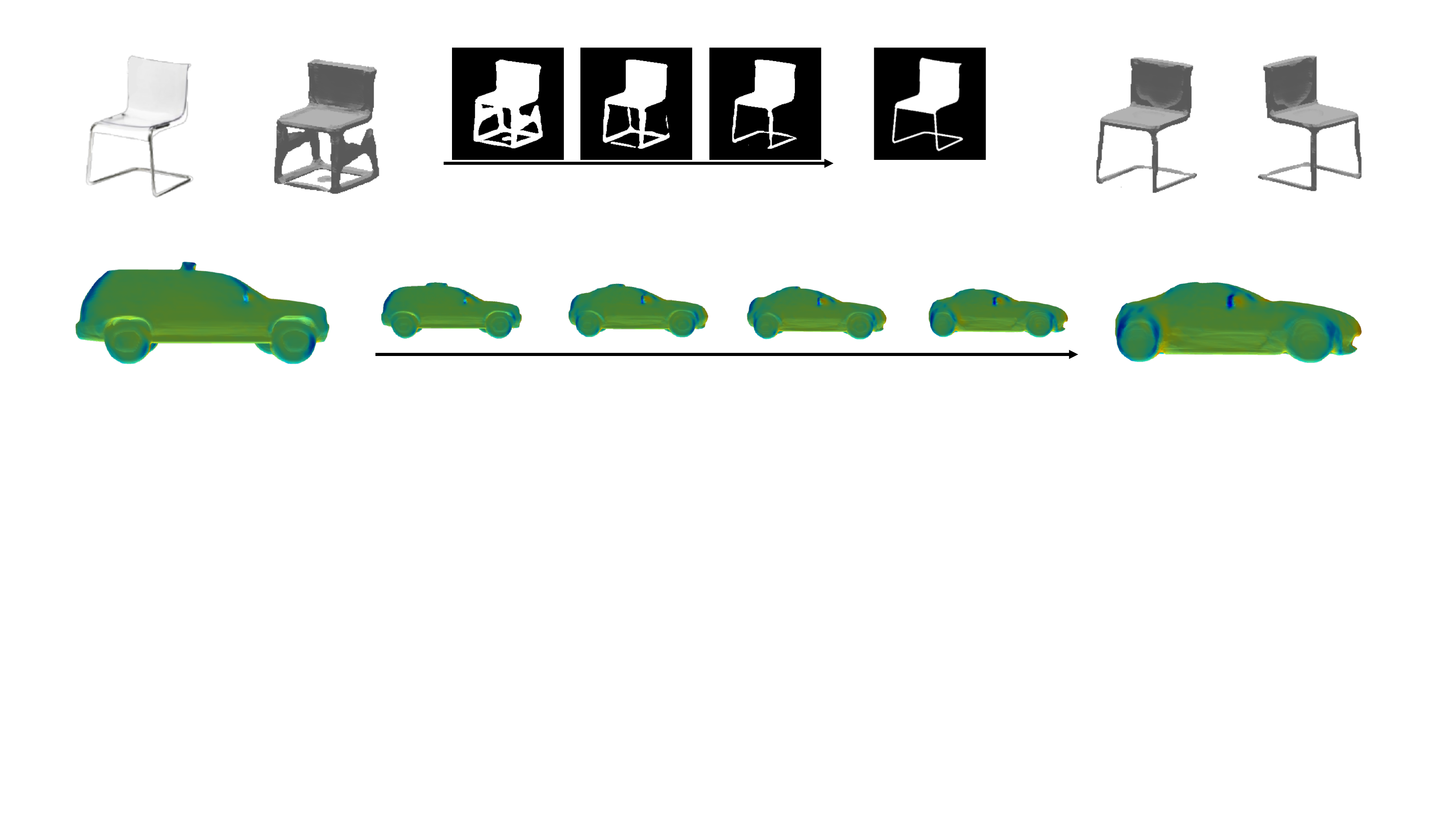}
			\put(6,25.5){\small{Image }}
			\put(16,25.5){\small{\textit{MeshSDF} Raw}}
			\put(35,25.5){\small{Silhouette refinement}}
			\put(60.5,26.5){\small{target}}
			\put(59,24.5){\small{silhouette}}
			\put(76,25.5){\small{\textit{MeshSDF} Refined}}
			
			\put(6,10.5){\small{\textit{MeshSDF} Raw}}
			\put(40,10.5){\small{Drag minimization}}
			\put(76,10.5){\small{\textit{MeshSDF} Refined}}
			
			\put(0,18.5){\small{(a) }}
			\put(0,4.5){\small{(b) }}
			\end{overpic}
		\end{center}
		\vspace{-10pt}
		\caption{\textbf{MeshSDF}. (a) We condition our representation on an input image and output an initial 3D mesh, which we refine via differentiable rasterization \cite{kato2018neural}, thereby exploiting MeshSDF's end-to-end differentiability. (b) We use our parameterization as a powerful regularizer for aerodynamic optimization tasks. Here, we start from an initial car shape and refine it to minimize pressure drag. }
\label{fig:teaser}
\vspace{-6pt}
\end{figure}

The non-differentiability of Marching Cubes has been addressed by learning differentiable approximations of it~\cite{Liao18a,Wickramasinghe20}. These techniques, however, remain limited to low-resolution meshes~\cite{Liao18a} or fixed topologies~\cite{Wickramasinghe20}. An alternative approach has been to reformulate downstream tasks, such as differentiable rendering~\cite{Jiang20a,Liu19NeurIPS} or surface reconstruction~\cite{Michalkiewicz19}, directly in terms of implicit functions, so that explicit surface representations are no longer needed. However, doing so is not easy and may even not be possible for more complex tasks, such as solving CFD optimization problems.

By contrast, we show that it is possible to use \textit{continuous} signed distance functions to produce explicit surface representations while preserving differentiability.  Our key insight is that 3D surface samples {\it can} be differentiated with respect to the underlying deep implicit field. We prove this formally by reasoning about how implicit field perturbations impact 3D surface geometry \textit{locally}. Specifically, we derive a closed-form expression for the derivative of a surface sample with respect to the underlying implicit field, which is independent of the method used to extract the iso-surface. This enables us to extract the explicit surface using a non-differentiable algorithm, such as Marching Cubes, and then perform our custom backward pass through the extracted surface samples, resulting in an end-to-end differentiable surface parameterization that can describe arbitrary topology and is not limited in resolution. We will refer to our approach as \textit{MeshSDF}.

We showcase the power and versatility of \textit{MeshSDF} in the two different applications depicted by Fig.~\ref{fig:teaser}. First, we exploit end-to-end differentiability to refine Single-View Reconstructions through differentiable surface rasterization~\cite{kato2018neural}. Second, we use our parameterization as powerful regularizer in physically-driven shape optimization for CFD purposes~\cite{Baque18}. We will demonstrate that in both cases our end-to-end differentiable parameterization gives us an edge over state-of-the art algorithms.

In short, our core contribution is a theoretically well-grounded technique for differentiating through iso-surface extraction.
This enables us to harness the full power of deep implicit surface representation to define an end-to-end differentiable surface mesh parameterization that allows topology changes.


%
\section{Related Work}
\label{sec:related}

\paragraph{From Discrete to Continuous Implicit Surface Models.}

Level sets of a 3D function effectively represent watertight surfaces with varying topology~\cite{Sethian99,Osher03}. As they can be represented on 3D grids and thus easily be processed by standard deep learning architectures, they have been an inspiration for many approaches~\cite{brock2016generative,Choy20163DR2N2AU,Gadelha16,Rezende16,Riegler_2017,Tatarchenko17,Wu20153DSA,Xie_2019}.
However, methods operating on dense grids have been limited to low resolution volumes due to excessive memory requirements. Methods operating on sparse representations of the grid tend to trade off the need for memory for a limited representation of fine details and lack of generalisation~\cite{Richter18, Riegler_2017,Tatarchenko17,Tatarchenko19}.

This has changed recently with the introduction of continuous deep implicit fields, which represent 3D shapes as level sets of deep networks that map 3D coordinates to a signed distance function~\cite{Park20a} or occupancy field~\cite{Mescheder19, Chen19c}. This yields a continuous shape representation wrt.~3D coordinates that is lightweight but not limited in resolution. This representation has been successfully used for single view reconstruction~\cite{Mescheder19,Chen19c,Xu19b} and 3D shape completion~\cite{Chibane20}. 

However, for applications requiring explicit surface parameterizations, the non-differentiability of iso-surface extraction so far has largely prevented exploiting the advantages of implicit representations.

\vspace{-6pt}
\paragraph{Converting Implicit Functions to Surface Meshes.}
The Marching Cube (MC) algorithm~\cite{Lorensen87,Newman06} is a widely adopted way
of converting implicit functions to surface meshes. The algorithm proceeds by sampling the field on a discrete 3D grid, detecting zero-crossing of the field along grid edges, and building a surface mesh using a lookup table. 
Unfortunately, the process of determining the position of vertices on grid edges involves linear interpolation, which does not allow for topology changes through backpropagation~\cite{Liao18a}, as illustrated in Fig.~\ref{fig:mc}(a). Because this is a central motivation to this work, we provide a more detailed analysis in the Supplementary Section. 

In what follows, we discuss two classes of methods that tackle the non-differentiability issue. The first one emulates iso-surface extraction with deep neural networks,
while the second one avoids the need for mesh representations by formulating objectives
directly in the implicit domain.
\vspace{-6pt}

\paragraph{\bf Emulating Iso-Surface Extraction.}
Liao et al.~\cite{Liao18a} map voxelized point clouds to a probabilistic topology distribution and vertex locations defined over a discrete 3D Euclidean grid through a 3D CNN. While this allows changes to surface topology through backpropagation, the probabilistic modelling requires keeping track of all possible topologies at the same time, which in practice limits resulting surfaces to low resolutions.
Voxel2mesh~\cite{Wickramasinghe20} deforms a mesh primitive and adaptively increases its resolution. While this enables high resolution surface meshes, it prevents changes of topology.
\vspace{-6pt}

\paragraph{\bf Reformulating Objective Functions in terms of Implicit Fields.}
In~\cite{michalkiewicz2019}, variational analysis is used to re-formulate standard surface mesh priors, such as those that enforce smoothness, in terms of implicit fields. Although elegant, this technique requires carrying out complex derivations for each new loss function and can only operate on an Euclidean grid of fixed resolution. 
The differentiable renderers of~\cite{jiang2020sdfdiff,liu2020dist} rely on sphere tracing and operate directly in terms of implicit fields.
Unfortunately, since it is computationally intractable to densely sample the underlying volume, these approaches either define implicit fields
over a low-resolution Euclidean grid \cite{jiang2020sdfdiff} or rely on heuristics to accelerate ray-tracing \cite{liu2020dist}, trading off in accuracy.
3D volume sampling efficiency can be improved by introducing a  sparse set of anchor points when performing ray-tracing~\cite{liu2019learning}. However, this requires reformulating standard surface mesh regularizers in terms of implicit fields using computationally intensive finite differences. Furthermore, these approaches~\cite{jiang2020sdfdiff,liu2019learning,liu2020dist} are tailored to differentiable rendering, and are not directly applicable to different settings that require explicit surface modeling, such as computational fluid dynamics.


\section{Method}
\label{sec:method}

Tasks such as Single-View Reconstruction (SVR)~\cite{Kanazawa18,Henderson19} or shape design in the context of CFD~\cite{Baque18} are commonly performed by deforming the shape of a 3D surface mesh $\mathcal{M}=(V,F)$, where $V = \{ \bv_1,\bv_2,... \}$ denotes vertex positions in $\mathbb R^3$ and $F$ facets, to minimize a task-specific loss function $\mathcal{L}_{\text{task}}(\mathcal{M})$. $\mathcal{L}_{\text{task}}$ can be, e.g., an image-based loss defined on the output of a differentiable renderer for SVR or a measure of aerodynamic performance for CFD. 

To perform surface mesh optimization robustly, a common practice is to rely on low-dimensional parameterizations that are either learned~\cite{Blanz99,Park20a,Bagautdinov18} or hand-crafted~\cite{Baque18,Umetani18,remelli17}. In that setting, a differentiable function maps a low-dimensional set of parameters $\mathbf z$ to vertex coordinates $V$, implying a fixed topology.

Allowing changes of topology, an implicit surface representation would pose a compelling alternative but conversely require a \textit{differentiable} conversion to explicit representations in order to backpropagate gradients of $\mathcal{L}_{\text{task}}$. 

In the remainder of this section, we first recapitulate deep Signed Distance Functions, which form the basis of our approach. We then introduce our main contribution, a differentiable approach to computing surface samples and updating their 3D coordinates to optimize $\mathcal{L}_{\textit{task}}$. Finally, we present \textit{MeshSDF}, a fully differentiable surface mesh parameterization that can represent arbitrary topologies.

\subsection{Deep Implicit Surface Representation}
\label{sec:rep}

We represent a generic watertight surface $S$ in terms of a \textit{signed distance function} (SDF) $s : \mathbb R ^ 3 \rightarrow \mathbb R$. Given the Euclidean distance $d(\bx,S)=\min_{\mathbf y \in {S}} d(\mathbf x, \mathbf y)$ of a 3d point $\bx$, $s(\bx)$ is $d(\bx,S)$ if $\bx$ is outside $S$ and $-d(\bx, {S})$ if it is inside. 
Given a dataset of watertight surfaces $\mathcal S$, such as ShapeNet~\cite{Chang15},  we train a Multi-Layer Perceptron $f_{\theta}$ as in~\cite{Park20a} to approximate $s$ over such set of surfaces $\mathcal S$  by minimizing
\begin{align}
\mathcal L_{\text{sdf}}(\{\mathbf z_S\}_{S \in \mathcal S}, \theta)  &= \sum _{S \in \mathcal S}\frac{1}{|X_S|}\sum _{\bx \in X_S}| f_\theta (\mathbf x , \bz_S) - s(\mathbf x) | + \lambda_{\text{reg}} \sum _{S \in \mathcal S} \| \bz_S \|_2^2 \; ,
\label{eq:sdf}
\end{align}
where $\bz_S \in \mathbb R ^ Z $ is the $Z$-dimensional encoding of surface $S$, $\theta$ denotes network parameters, $X_S$ represent 3D point samples we use to train our network and $\lambda_{\text{reg}}$ is a weight term balancing the contribution of reconstruction and regularization in the overall loss.

\subsection{Differentiable Iso-Surface Extraction}
\label{sec:diff}

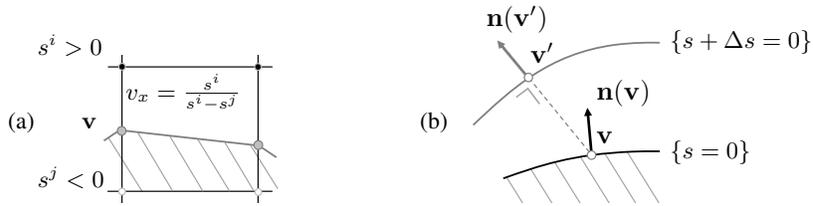
\begin{figure}[t]
\vspace{-6pt}
            \begin{center}
			\begin{overpic}[clip, trim=2.0cm 10cm 10cm 4cm,width=0.7\textwidth]{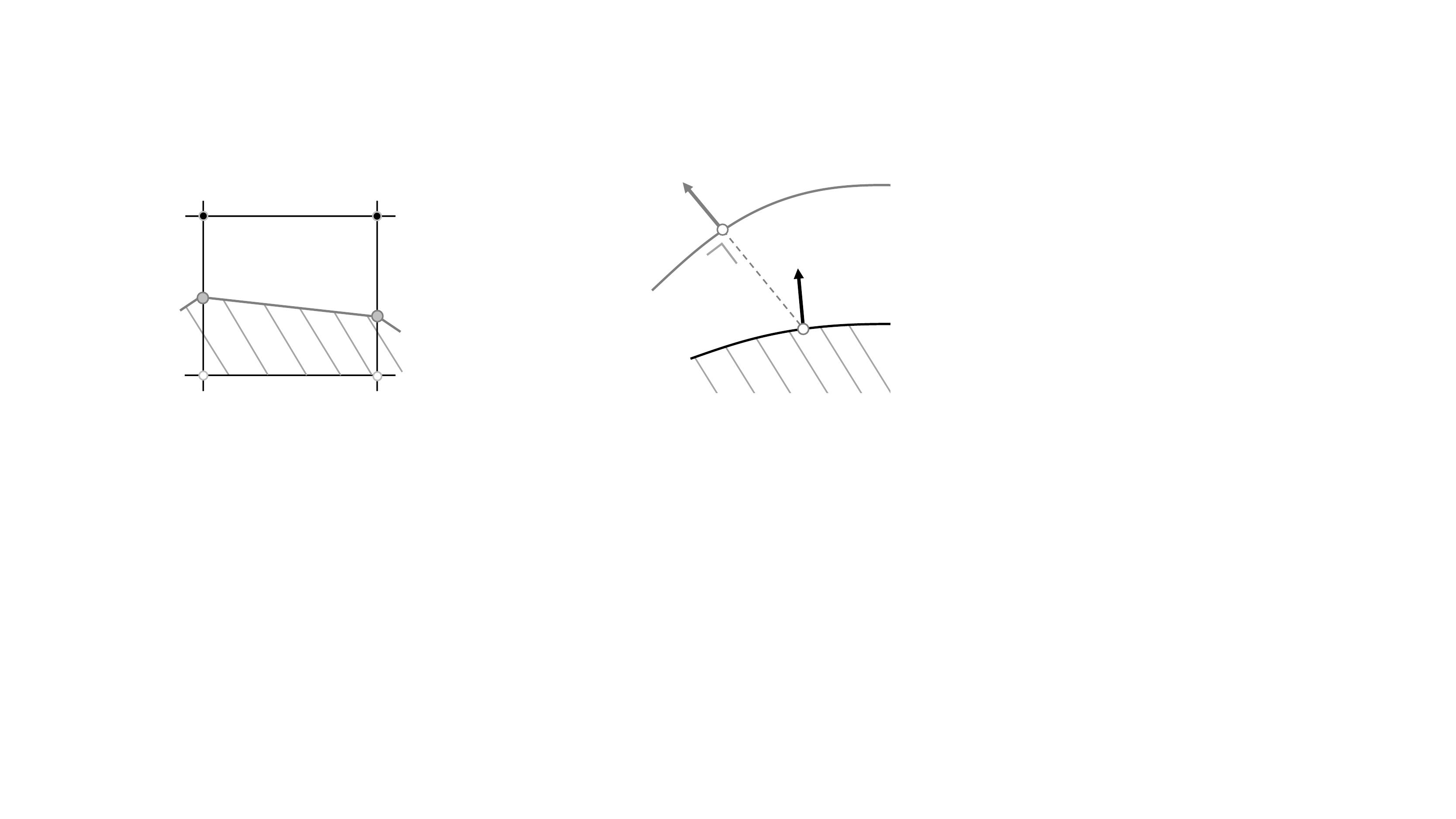}
			\put(1,20){\small{$ s ^i > 0 $}}
			\put(1, 2){\small{$ s ^j < 0 $}}
			\put(7,10){\small{$ \mathbf v $}}
			\put(13,14){\small{$ v_x = \frac{s ^i}{s ^i - s ^j } $}}
			
			\put(87,6){\small{$\{s = 0\} $}}
			\put(87,21){\small{$\{s +  \Delta s = 0\} $}}
			\put(77,8){$\mathbf v$}
			\put(77,14){$\mathbf n (\mathbf v)$}
			\put(68,19){$\mathbf v'$}
			\put(62,24){$\mathbf n (\mathbf v')$}
			
			\put(-3,10){\small{(a)}}
			\put(53,10){\small{(b)}}

			\end{overpic} 
			\end{center}
	\vspace{-3mm}
	\caption{\textbf{Marching cubes differentiation vs Iso-surface differentiation.} (a) Marching Cubes determines the position $v_x$ of a vertex $\bv$ along an edge via linear interpolation. This does not allow for effective back-propagation when topology changes because its behavior is degenerate when $s ^i=s ^j$ as shown in~\cite{Liao18a}. (b) Instead, we adopt a \textit{continuous} model expressed in terms of how signed distance function perturbations locally impact surface geometry. Here, we depict the geometric relation between local surface change $\Delta \mathbf v = \mathbf v' - \mathbf v$ and a signed distance perturbation  $\Delta s < 0$, which we exploit to compute $\frac{\partial \mathbf v}{\partial s}$ even when the topology changes.}
		\label{fig:mc}
\end{figure}

Once the weights $\theta$ of Eq.~\ref{eq:sdf} have been learned, $f_{\theta}$ maps a latent vector $\bz$ to a signed distance field and the surface of interest is its zero level set. Recall that our goal is to minimize the objective function $\mathcal{L}_{\text{task}}$ introduced at the beginning of this section.  As it takes as input a mesh defined in terms of its vertices and facets, evaluating it and its derivatives requires a \textit{differentiable} conversion from an implicit field to a set of vertices and facets, something that marching cubes does not provide, as depicted by Fig.~\ref{fig:mc}(a).
More formally, we need to be able to evaluate
\begin{align}
\frac{\partial \mathcal{L}_{\text{task}}}{\partial \bz} &= \sum_{{\bf v} \in V}\frac{\partial \mathcal{L}_{\text{task}}}{\partial {\bf v}}  \frac{\partial {\bf v}}{\partial f_{\theta}}\frac{\partial f_{\theta}}{\partial \bz} \; .
\label{eq:chainRule}
\end{align}

In this work, we take our inspiration from classical functional analysis~\cite{allaire2002level} and reason about the {\it continuous} zero-crossing of the SDF $s$ rather than focusing on how vertex coordinates depend on the implicit field $f_\theta$ when sampled by the marching cubes algorithm. This results in a differentiable approach to compute surface samples $\bv \in V$ from the underlying signed distance field $s$. We then simply exploit the fact that $f_\theta$ is trained to emulate a \textit{true} SDF $s$ to backpropagate gradients from $\mathcal{L}_{\text{task}}$ to the underlying deep implicit field $f_\theta$.

To this end, let us consider a generic SDF $s$, a point $\bv$ lying on its iso-surface $S = \{ \bq \in \mathbb{R}^3 | \;  s(\bq)=0 \}$, and see how the iso-surface moves when $s$ undergoes an infinitesimal perturbation $\Delta s$. Intuitively, $\Delta s <0$ yields a local surface inflation and $\Delta s >0$ a deflation, as shown in  Fig.~\ref{fig:mc}(b). In the Supplementary Section, we prove the following result, relating \textit{local} surface change $\Delta \mathbf v$ to field perturbation $\Delta s$.

\begin{theorem}
	\label{shape_derivative}
	Let us consider a signed distance function $s$ and a perturbation function $\Delta s$ such that $s+\Delta s$ is still a signed distance function. Given such $\Delta s$, we define the associated local surface change $\Delta \mathbf v = \mathbf v' - \mathbf v$ as the displacement between  $ \mathbf v'$, the closest point to  surface sample $\mathbf v $ on the perturbed surface $S' = \{ \bq \in \mathbb{R}^3 | \;  s + \Delta s (\bq) =0 \}$, and the original surface sample $\mathbf v$.
	It then holds that
	\begin{align}
	\frac{\partial \mathbf v}{\partial s } (\mathbf v) = - \mathbf n (\mathbf v) = -\nabla s (\mathbf v) \; , 
	\end{align}
	where $\mathbf n$ denotes the surface normals.
\end{theorem}

Because $f_\theta$ is trained to closely approximate a signed distance function $s$, we can now replace $\frac{\partial {\bf v}}{\partial f_{\theta}}$ in 
Eq.~\ref{eq:chainRule} by $-\nabla f_\theta (\bv,\bz)$, which yields
\begin{align}
\label{eq:backward}
\frac{\partial \mathcal{L}_{\text{task}}}{\partial \bz} = \sum_{\mathbf v \in V} -\frac{\partial \mathcal{L}_{\text{task}}}{\partial \mathbf v}  \nabla f_\theta (\bv,\bz)   \frac{ \partial f_{\theta} }{ \partial \bz}(\bv,\bz) \; .
\end{align}

In short, given an objective function defined with respect to surface samples $\bv \in V$, we can back-propagate gradients all the way back to the latent code $\bz$, which means that we can define a mesh representation that is differentiable end-to-end while being able to capture changing topologies, as will be demonstrated in Section~\ref{sec:exp}. 

When performing a forward pass, we simply evaluate our deep signed distance field $f_\theta$ on an Euclidean grid, and use marching cubes (MC) to perform iso-surface extraction and obtain surface mesh $\mathcal M = (V,F)$. Conversely, we follow the chain rule of Eq.~\ref{eq:backward} to assemble our backward pass. This requires us to perform an additional forward pass of surface samples $\bv \in V$ to compute surface normals $\nabla f_\theta (\bv)$ as well as $\frac{ \partial f_{\theta} }{ \partial \bz}(\bv,\bz)$. We implement \textit{MeshSDF} following the steps detailed in Algorithms~\ref{algo:mesh-fwd} and~\ref{algo:mesh-back}.
Refer to the Supplementary Section for a detailed analysis of the computational burden of iso-surface extraction within our pipeline.

\vspace{-3pt}
\begin{minipage}[t]{0.48\textwidth}
\begin{algorithm}[H]
    \centering
    \caption{MeshSDF Forward}\label{algo:mesh-fwd}
    \begin{algorithmic}[1]
        \State \text{\textbf{input:} latent code $\bz$} 
        \State \text{\textbf{output:} surface mesh $\mathcal M = (V,F)$} 
        
        \State \text{assemble grid $G_{3D}$} 
        \State \text{sample field on grid $S = f_\theta(\bz, G_{3D})$} 
        \State \text{extract iso-surface $(V,F) = $ MC$(S, G_{3D})$} 
        \State \textbf{Return}  $\mathcal M = (V,F)$
    \end{algorithmic}
\end{algorithm}
\end{minipage}
\hfill
\begin{minipage}[t]{0.48\textwidth}
\begin{algorithm}[H]
    \centering
    \caption{MeshSDF Backward}\label{algo:mesh-back}
    \begin{algorithmic}[1]
        \State \text{\textbf{input:} upstream gradient $\frac{\partial \mathcal L}{\partial \bv }$ for $\bv \in V$} 
        \State \text{\textbf{output:} downstream gradient $\frac{\partial \mathcal L}{\partial \bz}$} 
        \State \text{forward pass $s_{\bv} = f_\theta(\bz, \bv)$ for $\bv \in V$} 
        \State \text{$\mathbf n(\bv)  = \nabla f_\theta(\bz, \bv)$  for $\bv \in V$} 
        \State \text{$\frac{\partial \mathcal L}{\partial f_\theta}(\bv) = - \frac{\partial  \mathcal L}{\partial \bv} \; \mathbf n $ for $\bv \in V$} 
        \State \textbf{Return}  $\frac{\partial \mathcal L}{\partial \bz} = \sum_{\mathbf v \in V} \frac{\partial \mathcal L}{\partial f_\theta}(\bv) \frac{\partial f_\theta}{\partial \bz}(\bv)$
    \end{algorithmic}
\end{algorithm}
\end{minipage}

\begin{figure}[t]
		\vspace{-10pt}
		\hspace{-2pt}
		\begin{center}
			\begin{overpic}[clip, trim=2.2cm 7.0cm 2.2cm 6.5cm,width=1.0\textwidth]{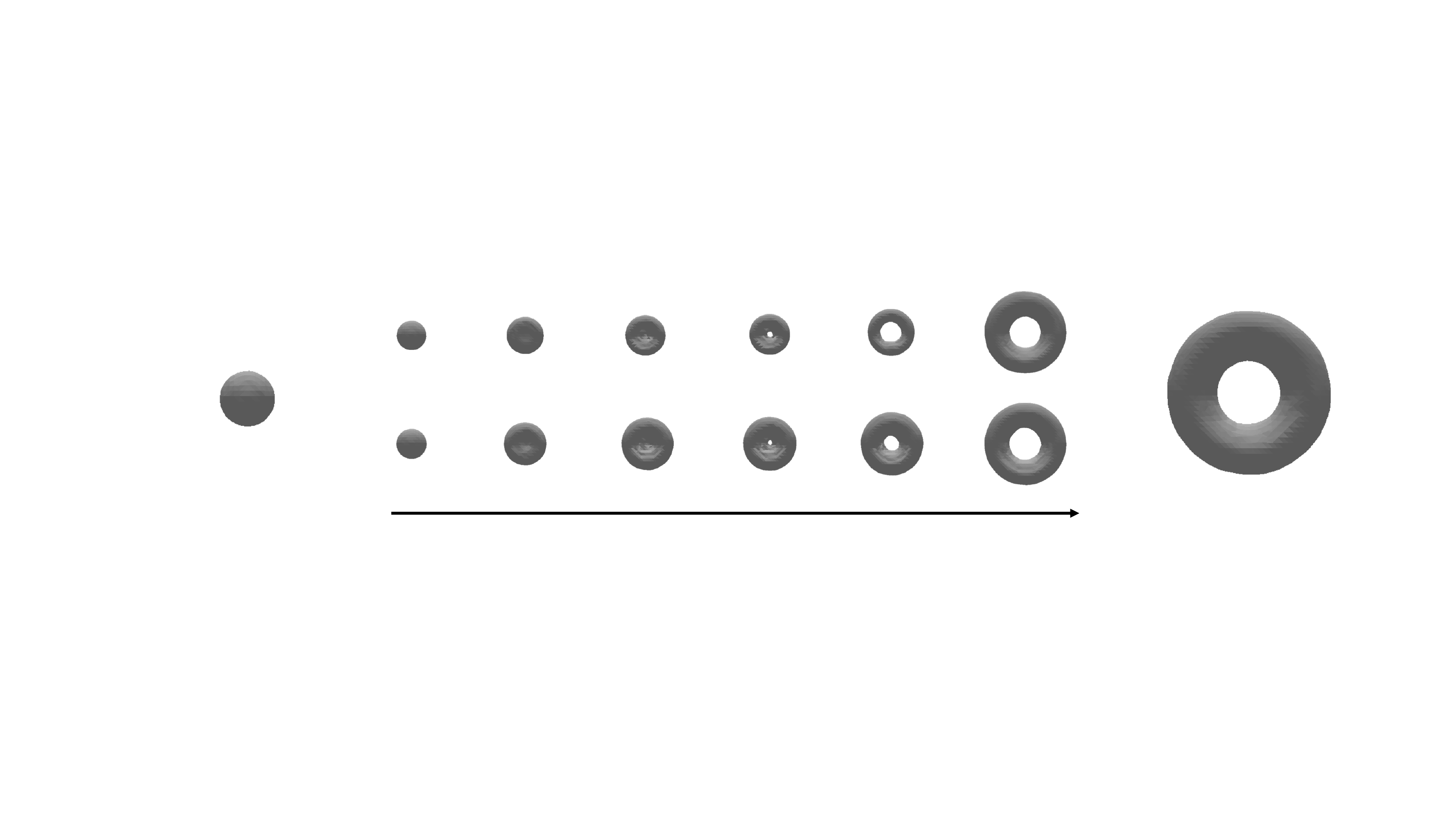}
			\put(40,1){\small{optimization iterations}}
			
			\put(38,19){\small{(a) surface-to-surface distance }}
			\put(39,10.1){\small{(b) image-to-image distance }}
			
			\put(6,19){{initialization }}
			\put(88,19){{target}}
			
			\put(11,1){{$S$}}
			\put(90,1){{$T$}}
			
			\end{overpic}
		\end{center}
		\vspace{-10pt}
		\caption{\textbf{Topology-Variant Parameterization.}
 We minimize (a) a surface-to-surface or (b) an image-to-image distance with respect to the latent vector $\bz$ to transform a sphere (genus-0) into a torus (genus-1).
 This demonstrates that we can backpropagate gradient information from mesh vertices to latent vector while modifying surface mesh topology.}
\label{fig:sanity}
\vspace{-6pt}
\end{figure}
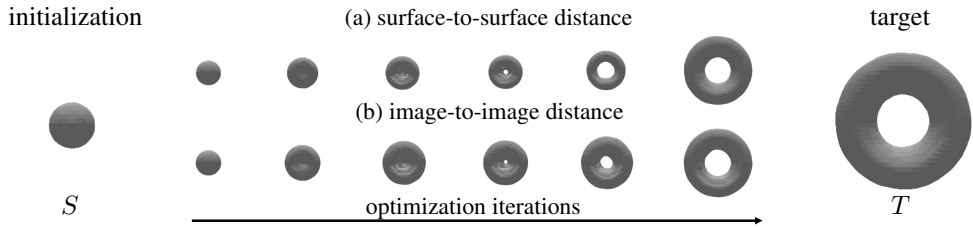

\section{Experiments}
\label{sec:exp}

We first use a simple example to show that, unlike marching cubes, our approach allows for differentiable topology changes. We then demonstrate that we can exploit surface mesh differentiability to outperform state-of-the-art approaches on two very different tasks, Single View Reconstruction\footnote[1]{main corresponding author: edoardo.remelli@epfl.ch} and Aerodynamic Shape Optimization\footnote[2]{main corresponding author: artem.lukoianov@epfl.ch}. 

\subsection{Differentiable Topology Changes}

In the experiment depicted by Fig.~\ref{fig:sanity}, we used a database of spheres and tori of varying radii to train a network $f_\theta$ that implements the approximate signed function $s$ of Eq.~\ref{eq:sdf}. As a result, $f_\theta$ associates to a latent vector $\bz$ an implicit field $f_\theta(\bz)$ that defines spheres, tori, or a mix of the two. 

We now consider two loss functions that operate on explicit surfaces $S$ and $T$
\begin{align}
    \mathcal{L}_{\text{task1}}  &= \min _{s \in S} d(s,T) + \min _{t \in T} d(S,t) \; , \\
    \mathcal{L}_{\text{task2}}   &=\| \text{DR}(S)- \text{DR}(T)\|_1 \; , 
\end{align}
where $d$ is the point-to-surface distance in 3D \cite{ravi2020pytorch3d} and $\text{DR}$ is the output of an off-the-shelf differentiable rasterizer~\cite{kato2018neural}, that is $\mathcal{L}_{\text{task1}}$ is the surface-to-surface distance while $\mathcal{L}_{\text{task2}}$ is the image-to-image distance between the two rendered surfaces.

In the example shown in Fig.~\ref{fig:sanity}, $S$ is the sphere on the left and $T$ is the torus on right. We initialize the latent vector $\bz$ so that it represents $S$. We then use the pipeline of Sec.~\ref{sec:diff} to minimize either $\mathcal{L}_{\text{task1}}$ or $\mathcal{L}_{\text{task2}}$, backpropagating surface gradients to the underlying implicit representation. 
In both cases, the sphere smoothly turns into a torus, thus changing its genus. 
Note that even though we rely on a deep signed distance function to represent our topology-changing surfaces, we did {\it not} have to reformulate the loss functions in terms of implicit surfaces, as done in ~\cite{michalkiewicz2019,jiang2020sdfdiff,liu2020dist,liu2019learning}. 
We now turn to demonstrating the benefits of having a topology-variant surface mesh representation through two concrete applications, Single-View Reconstruction and Aerodynamic Shape Optimization.

\subsection{Single-View Reconstruction}

Single-View Reconstruction (SVR) has emerged as a standardized benchmark to evaluate 3D shape representations \cite{Choy20163DR2N2AU,fan17,groueix2018papier,Wang_2018,Chen19c,Mescheder19,Pontes2018Image2MeshAL,Gkioxari_2019,Richter18,xu2019disn,Tatarchenko19}. 
We demonstrate that our method is straightforward to apply to this task and validate our approach on two standard datasets, namely ShapeNet~\cite{Chang15} and Pix3D~\cite{Sun2018Pix3DDA}. More results, as well as failure cases, can be found in the Supplementary material.

\paragraph{Differentiable Meshes for SVR.}

As in~\cite{Mescheder19,Chen19c}, we condition our deep implicit field architecture on the input images via a residual image encoder~\cite{He_2016}, which maps input images to latent code vectors $\bz$. These latent codes are then used to condition the architecture of Sec.~\ref{sec:rep} and compute the value of deep implicit function $f_{\theta}$. Finally, we minimize $\mathcal L_{\text{sdf}}$ (Eq.~\ref{eq:sdf}) wrt. $\theta$ on a training set of image-surface pairs. 
This setup forms our baseline approach, \emph{MeshSDF} (raw).

To demonstrate the effectiveness of the surface representation proposed in Sec.~\ref{sec:diff}, we exploit differentiability during inference via differentiable rasterization~\cite{kato2018neural}. We refer to this variant as \emph{MeshSDF}\@.
Similarly to our baseline, during inference, the encoder predicts an initial latent code $\bz$. Different to our baseline, our full version refines the predicted shape $\mathcal M$, as depicted by the top row of Fig.~\ref{fig:teaser}.
That is, given the camera pose associated to the image and the current value of $\bz$, we project vertices and facets into a binary silhouette in image space through a differentiable rasterization function $\text{DR}_{\text{silhouette}}$~\cite{kato2018neural}. Ideally, the projection matches the observed object silhouette $\cal S$ in the image, which is why we define our objective function as
\begin{equation}
\mathcal{L}_{\text{task}} = \| \text{DR}_{\text{silhouette}}(\mathcal M(\bz))- \mathcal S\|_1 \:, 
\end{equation}
which we minimize with respect to $\bz$. In practice, we run 400 gradient descent iterations using Adam~\cite{kingma2014adam} and keep the $\bz$ with the smallest $\mathcal{L}_{\text{task}}$ as our final code vector. 

\paragraph{Comparative results on ShapeNet.}
We report our results on ShapeNet~\cite{chang2015shapenet} in Tab.~\ref{tab:shapenet}. 
We compare our approach against state-of-the-art mesh reconstruction approaches: reconstructing surface patches~\cite{groueix2018papier}, generating surface meshes with fixed topology~\cite{Wang_2018}, generating meshes from voxelized intermediate representations~\cite{Gkioxari_2019}, and representing surface meshes using signed distance functions~\cite{xu2019disn}.
We used standard train/test splits along with the renderings provided in~\cite{xu2019disn} for all the methods we tested. We evaluate on standard SVR metrics~\cite{Tatarchenko19}, which we define in the Supplementary Section. 
We report our results in Tab.~\ref{tab:shapenet}. {\it MeshSDF} (raw) refers to reconstructions using our encoder-decoder architecture, which is similar to those of~\cite{Mescheder19,Chen19c}, without any further refinement. Our full method, {\it MeshSDF}, exploits end-to-end differentiability to minimize $\mathcal{L}_{\text{task}}$  with respect to $\bz$. This improves performance by at least $12\%$ over {\it MeshSDF} (raw) on all metrics. As a result, our full approach also outperforms all other state-of-the-art approaches. 

\begin{table*}[ht]
    \caption{\textbf{Single view reconstruction results on ShapeNet Core.} Exploiting end-to-end differentiability to perform image-based refinement allows us to outperform all prior methods.}
	\label{tab:shapenet}
	\begin{center}
		\scalebox{0.65}{
			\begin{tabular}{c|c|ccccccccccccc|c}
				\Xhline{2\arrayrulewidth}
				Metric & Method & plane & bench & cabinet & car & chair & display & lamp & speaker & rifle & sofa & table & phone & boat & mean \\
				\Xhline{2\arrayrulewidth}
				\multirow{6}{*}{IoU $\uparrow$}&AtlasNet~\cite{groueix2018papier}&20&13&7&16&13&12&14&8&28&11&15&14&17&15\\
                &Mesh R-CNN~\cite{Gkioxari_2019}&24&25&17&21&21&21&20&15&32&19&26&26&26&23 \\
                &Pixel2Mesh~\cite{Wang_2018}&29&32&\textbf{22}&25&27&27&\textbf{28}&19&40&23&\textbf{31}&36&32&29 \\
                &DISN~\cite{xu2019disn}&\textbf{40}&33&20&31&25&33&21&19&\textbf{60}&\textbf{29}&25&44&\textbf{34}&30 \\
				\cline{2-16}
				&MeshSDF (raw) &32&32&19&30&24&28&20&18&45&26&24&48&28&28\\
				&MeshSDF &36&\textbf{38}&\textbf{22}&\textbf{32}&\textbf{28}&\textbf{34}&25&\textbf{22}&52&\textbf{29}&\textbf{31}&\textbf{54}&30&\textbf{32}\\
				\Xhline{2\arrayrulewidth}
				\multirow{6}{*}{EMD $\cdot 10^2$ $\downarrow$}&AtlasNett~\cite{groueix2018papier}&6.3&7.9&9.5&8.3&7.8&8.8&9.8&10.2&6.6&8.2&7.8&9.9&7.1&8.0\\
                &Mesh R-CNN~\cite{Gkioxari_2019}&4.5&3.7&4.3&3.8&4.0&4.6&5.7&5.1&3.8&4.0&3.9&4.7&4.1&4.2 \\
                &Pixel2Mesh~\cite{Wang_2018}&3.8&2.9&3.6&3.1&3.4&3.3&4.8&3.8&3.2&3.1&3.3&2.8&3.2&3.4 \\
                &DISN~\cite{xu2019disn}&\textbf{2.2}&2.3&3.2&2.4&2.8&2.5&3.9&3.1&\textbf{1.9}&\textbf{2.3}&2.9&1.9&\textbf{2.3}&2.6 \\
				\cline{2-16}
				&MeshSDF (raw) &3.3&2.5&3.2&2.2&2.8&3.0&4.2&3.5&2.6&2.7&3.1&1.9&2.9&3.0\\
				&MeshSDF &2.5&\textbf{2.1}&\textbf{3.0}&\textbf{2.0}&\textbf{2.4}&\textbf{2.4}&\textbf{3.2}&\textbf{2.9}&\textbf{1.9}&2.4&\textbf{2.7}&\textbf{1.7}&\textbf{2.3}&\textbf{2.5}\\
				\Xhline{2\arrayrulewidth}
				\multirow{6}{*}{CD-$l_2 \cdot 10^3$ $\downarrow$}&AtlasNett~\cite{groueix2018papier}&10.6&15.0&30.7&10.0&11.6&17.3&17.0&22.0&6.4&11.9&12.3&12.2&10.7&13.0\\
                &Mesh R-CNN~\cite{Gkioxari_2019}&13.3&8.3&10.5&7.2&9.8&10.9&16.4&14.8&6.9&8.7&10.0&6.9&10.4&10.3 \\
                &Pixel2Mesh~\cite{Wang_2018}&12.4&5.5&8.2&5.6&6.9&8.2&\textbf{12.3}&\textbf{11.2}&6.0&6.8&\textbf{7.9}&4.7&7.9&8.0 \\
                &DISN~\cite{xu2019disn}&\textbf{6.3}&6.6&11.3&5.3&9.6&8.6&23.6&14.5&4.4&\textbf{6.0}&12.5&5.2&\textbf{7.8}&9.7\\
				\cline{2-16}
				&MeshSDF (raw) &10.6&9.5&8.8&4.2&8.2&12.4&25.9&20.4&8.9&11.5&14.6&6.2&17.1&12.0\\
				&MeshSDF &\textbf{6.3}&\textbf{5.4}&\textbf{7.8}&\textbf{3.5}&\textbf{5.9}&\textbf{7.3}&14.9&12.1&\textbf{3.4}&7.8&10.7&\textbf{3.9}&10.0&\textbf{7.8}\\
				\Xhline{2\arrayrulewidth}
		\end{tabular}}
	\end{center}
\end{table*}

\paragraph{Comparative results on Pix3D.}

Whereas ShapeNet contains only rendered images, Pix3D~\cite{Sun2018Pix3DDA} is a test dataset that comprises \comment{10069}real images paired to \comment{395 unique} 3D models. We follow the evaluation protocol and metrics proposed in ~\cite{Sun2018Pix3DDA}, which we detail in the supplementary material.

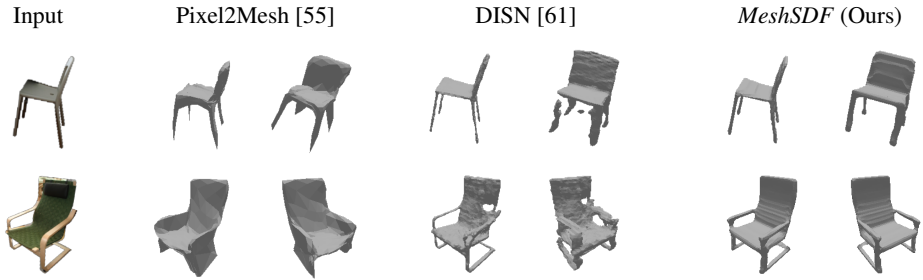
\begin{figure}[t]
		\begin{center}
			\begin{overpic}[clip, trim=0.0cm 10cm 0cm 0.5cm,width=1.0\textwidth]{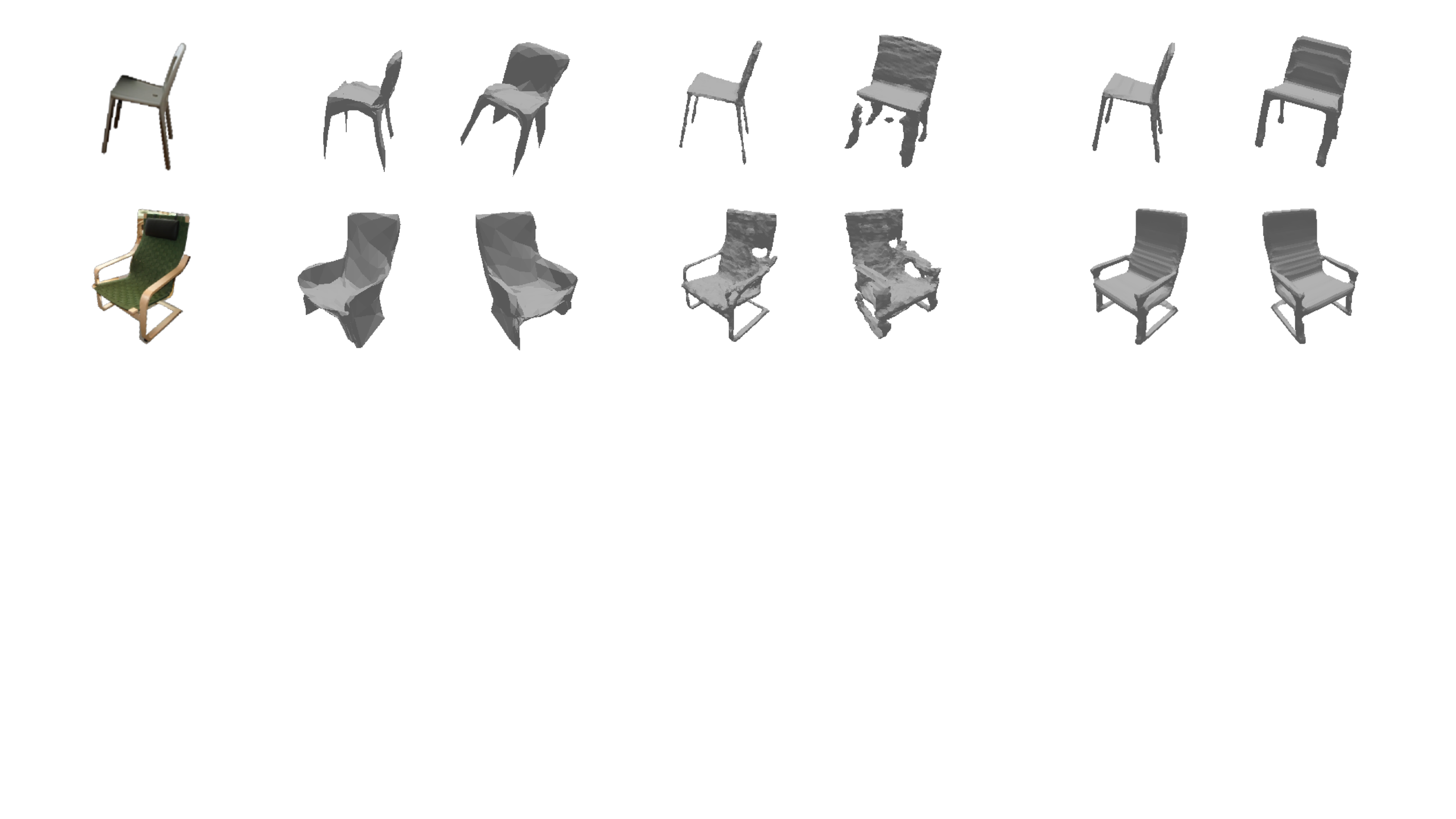}
			\put(7,27){\small{Input }}
			\put(22.5,27){\small{Pixel2Mesh} \cite{Wang_2018}}
			\put(51,27){\small{DISN} \cite{xu2019disn}}
			\put(76,27){\small{\textit{MeshSDF} (Ours) }}

			\end{overpic}
		\end{center}
		\vspace{-14pt}
		\caption{\textbf{Pix3D Reconstructions.} We compare our refined predictions to the runner-up approaches for the experiment in Tab. \ref{tab:pix3d}. \textit{MeshSDF} can represent arbitrary topology as well as learn strong shape priors, resulting in reconstructions that are consistent even when observed from view-points different from the input one. }
\label{fig:pix3D}
\vspace{-3pt}
\end{figure}

For this experiment we use the same function $f_\theta$ as for ShapeNet, that is, we do not fine-tune our model on Pix3D images, but train it on synthetic chair renders only so that to encourage the learning of stronger shape priors.
We report our results in Tab.~\ref{tab:pix3d} and in Fig.~\ref{fig:pix3D}. 
Interestingly, in this more challenging setting using real-world images, our simple baseline \textit{MeshSDF} (raw) already performs on par with more sophisticated methods using camera information \cite{xu2019disn}. As for ShapeNet, our full model outperforms all other approaches.

\begin{table*}[ht]
	\centering
	\small
	\caption{\textbf{Single view reconstruction results on Pix3D Chairs.} Our full approach outperforms all prior methods in all metrics.}
	\label{tab:pix3d}
	\begin{tabularx}{\linewidth}{@{}X*5{@{\hspace{2mm}}c}@{\hspace{3mm}}c@{\hspace{2mm}}c@{}}
		\toprule
		\myfootnotesize Metric & \myfootnotesize Pix3D \cite{Sun2018Pix3DDA} & \myfootnotesize AtlasNet \cite{groueix2018papier} & \myfootnotesize Mesh R-CNN \cite{Gkioxari_2019} & \myfootnotesize Pixel2Mesh \cite{Wang_2018} & \myfootnotesize DISN \cite{xu2019disn} & \myfootnotesize MeshSDF (raw) & \myfootnotesize MeshSDF\\
		\midrule
		IoU $\uparrow$  & 0.282  & {-} &  0.240 & 0.254 &  0.333 & 0.337 & \textbf{0.407}\\
		EMD $\downarrow$ & 0.118  & 0.128 &  0.125 & 0.115 & 0.117 & 0.119 & \textbf{0.098}\\
		CD-$\sqrt{l_2}$ $\downarrow$ & 0.119  & 0.125  &  0.110 & 0.104 & 0.104 & 0.102 & \textbf{0.089}\\
		\bottomrule
	\end{tabularx}
\end{table*}

\subsection{Shape Optimization}

Computational Fluid Dynamics (CFD)  plays a central role in designing cars, airplanes and many other machines. It typically involves approximating the solution of the Navier-Stokes equations using numerical methods. Because this is computationally demanding, \textit{surrogate} methods~\cite{Toal11,Xu17,Baque18,Umetani18} have been developed to infer physically relevant quantities, such as pressure field, drag or lift, directly from 3D surface meshes without performing actual physical simulations. This makes it possible to optimize these quantities with respect to the 3D shape using gradient-based methods and at a much lower computational cost. 

In practice, the space of all possible shapes is immense. Therefore, for the optimization to work well, one has to parameterize the space of possible shape deformations, which acts as a strong regularizer. In~\cite{Baque18,Umetani18} hand-crafted surface parameterizations were introduced. It was effective but not generic and had the potential to significantly restrict the space of possible designs. We show here that we can use \textit{MeshSDF} to improve upon hand-crafted parameterizations.

\paragraph{Experimental Setup.}

We started with the ShapeNet car split by automatic deletion of all the internal car parts \cite{sin2013vega} and then manually selected $N=1400$ shapes suitable for CFD simulation. For each surface ${\mathcal M _i}$ we ran OpenFoam~\cite{jasak2007openfoam} to predict a pressure field $p_i$ exerted by air travelling at 15 meters per second towards the car. The resulting training set  $\{ \mathcal M _i, p_i \}_{i=1}^N$ was then used to train a Mesh Convolutional Neural Network~\cite{Fey/etal/2018} $g_{\beta}$ to predict the pressure field $p=g_{\beta}(\mathcal M)$, as in~\cite{Baque18}. We use $\{\mathcal M_i\}_{i=1}^N$ to also learn the representation of Sec.~\ref{sec:diff} and train the network that implements $f_{\theta}$ of Eq.~\ref{eq:sdf}.

Finally, we introduce the aerodynamic objective function
\begin{align}
\label{eq:cfd}
    \mathcal L _{\text{task}}(\mathcal{M}) = \iint _{\mathcal{M}} \, g_\beta \, \mathbf n_x \, \text{d}\mathcal{M} + \mathcal L _{\text{constraint}}(\mathcal{M}) \; ,
\end{align}
where the integral term approximates drag given the predicted pressure field, $\mathbf n_x$ denotes the projection of surface normals along airflow direction, and $\mathcal{L}_{\text{constraint}}$ is designed to preserve the required amount of space for the engine and the passenger compartment. Minimizing the drag of the car can now be achieved by minimizing $\mathcal L _{\text{task}}$ with respect to $\mathcal M$. We provide further details about this process and the justification for our definition of $\mathcal L _{\text{task}}$ in the Supplementary Section. 

\paragraph{Comparative Results.}

We compare our surface parameterization to several baselines: (1) vertex-wise optimization, that is, optimizing the objective with respect to each vertex; (2) scaling the surface along its 3 principal axis; (3) using the \textit{FreeForm} parameterization of~\cite{Baque18}, which extends scaling to higher order terms as well as periodical ones and (4) the \textit{PolyCube} parameterization of~\cite{Umetani18} that deforms a 3D surface by moving a pre-defined set of control points. 

\begin{figure}[t]
		\vspace{-10pt}
		\begin{center}
			\begin{overpic}[clip, trim=2.5cm 3cm 0cm 3cm,      
			                        width=\textwidth]{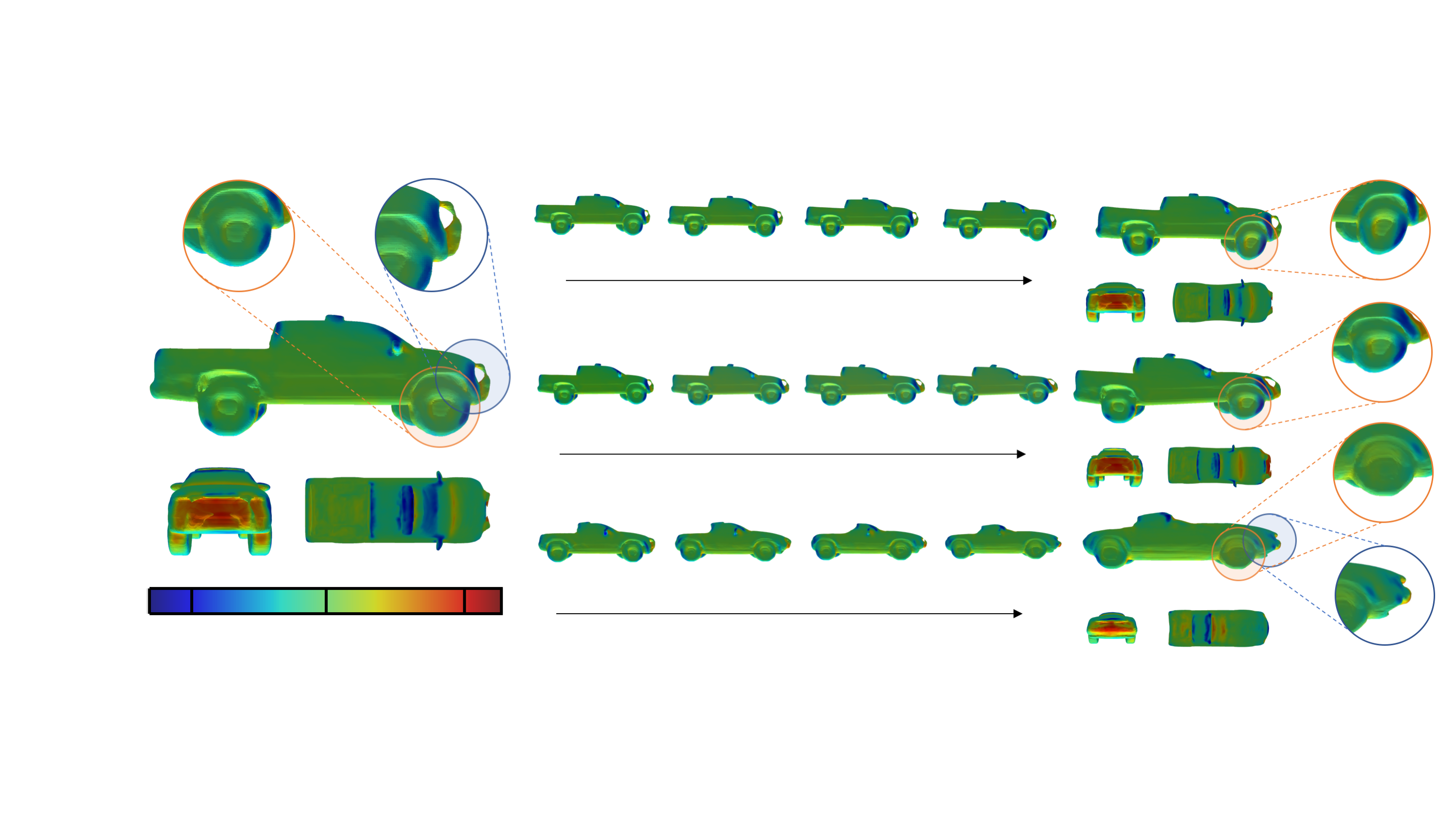}
    			\put(40,6){\small{{MeshSDF}}}
    			\put(40,17.83){\small{PolyCube }}
    			\put(40,30.75){\small{FreeForm }}
    			
    			\put(25, 3.5){\small{$p_{max}$}}
    			\put(15.5, 3.5){\small{$0$}}
    			\put(5, 3.5){\small{$p_{min}$}}
    			
    			\put(72, 41){\small{optimized shape}}
    			\put(10, 41){\small{initial shape}}
    			
    			\put(55,6){\small{$0.597$}}
    			\put(55,17.83){\small{$0.852$}}
    			\put(55,30.75){\small{$0.889$}}
    			
			\end{overpic}
		\end{center}
\vspace{-4mm}
\caption{\textbf{Drag minimization.} Starting from an initial shape (left column), $\mathcal L _{\textit{task}}$ is minimized  using three different parameterizations: FreeForm (top), PolyCube (middle), and our \textit{MeshSDF} (bottom). The middle column depicts the optimization process and the relative improvements in terms of $\mathcal L _{\textit{task}}$. The final result is shown in the right column.
FreeForm and PolyCube lack a semantic prior, resulting in implausible details such as sheared wheels (orange inset). By contrast, \textit{MeshSDF} not only enforces such priors but can also effect topology changes (blue inset).}
\label{fig:cfd_three_baselines}
\end{figure}
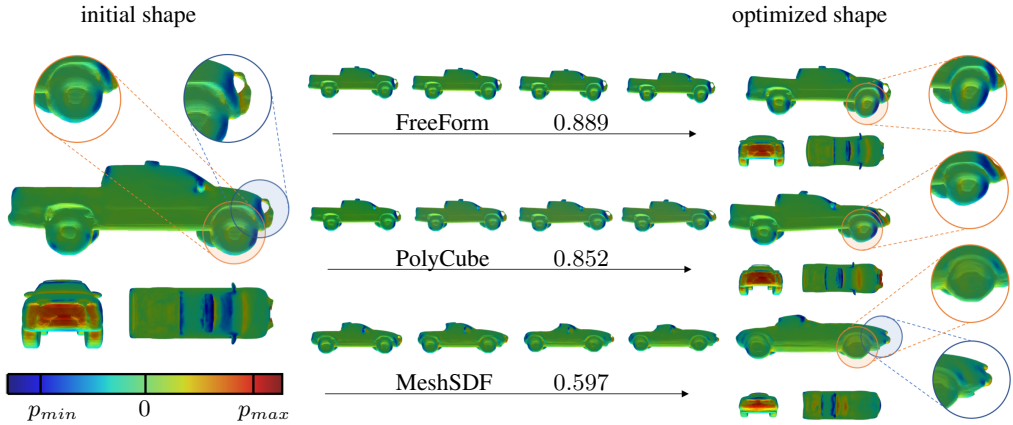

We report quantitative results for the minimization of the objective function of Eq.~\ref{eq:cfd} for a subset of 8 randomly chosen cars in Table~\ref{tab:cfd_results}, and show qualitative ones in Fig.~\ref{fig:cfd_three_baselines}. Not only does our method deliver lower drag values than the others but, unlike them, it allows for topology changes and produces semantically correct surfaces as shown in Fig.~\ref{fig:cfd_three_baselines}(c). 

\begin{table*}[h]
	\centering
	\small
	\caption{\textbf{CFD-driven optimization}.We minimize drag on car shapes comparing different surface parameterizations. Numbers in the table (avg $\pm$ std) denote {relative improvement of the objective function $\mathcal L _{\text{task}}^{\%} = \mathcal L _{\text{task}} / \mathcal L _{\text{task}}^{t=0} $ } for the optimized shape, as obtained by CFD simulation in OpenFoam.}
	\label{tab:cfd_results}
	\begin{tabularx}{\linewidth}{@{}X*4{@{\hspace{2mm}}c}@{\hspace{3mm}}c@{\hspace{2mm}}c@{}}
		\toprule
		\myfootnotesize Parameterization & \myfootnotesize None & \myfootnotesize Scaling & \myfootnotesize FreeForm \cite{Baque18}  & \myfootnotesize PolyCube \cite{Umetani18} & \myfootnotesize MeshSDF\\
		\midrule
		Degrees of Freedom  & $\sim100$k &  $3$  & $21$  &  $\sim 332$ & $256$\\
		Simulated {$\mathcal L _{\text{task}}^{\%} $} $\downarrow$ 
		& not converged  & $0.931 \pm 0.014$ &  $0.844 \pm 0.171$ & $0.841 \pm 0.203$ & $\textbf{0.675} \pm \textbf{0.167}$\\
		\bottomrule
	\end{tabularx}
\end{table*}

\section{Conclusion}

We introduce a new approach to extracting 3D surface meshes from Deep Signed Distance Functions while preserving end-to-end differentiability. This enables combining powerful implicit models with objective functions requiring explicit representations such as surface meshes. 
We believe that \textit{MeshSDF} will become particularly useful for Computer Assisted Design, where having a topology-variant explicit surface parameterizations opens the door to new applications.

\section{Acknowledgments}
This project was supported in part by the Swiss National Science Foundation. 

\section{Broader Impact}

Computational Fluid Dynamics is key to addressing the critical engineering problem of designing shapes that maximize aerodynamic, hydrodynamic, and heat transfer performance, and much else beside. The techniques we propose therefore have the potential to have a major impact in the field of Computer Assisted Design by unleashing the full power of deep learning in an area where it is not yet fully established.

{\small
\bibliographystyle{ieee_fullname}
\bibliography{  bib/string,
                bib/vision,
                bib/cfd,
                bib/biomed,
                bib/graphics,
                bib/learning,
                bib/egbib}
}

\newpage
\section{Supplementary Material}

In this supplementary material, we first remind the interested reader of why marching cubes are not differentiable and provide a formal proof of our main differentiability theorem. We then discuss our approach to speeding-up iso-surface extraction and performing end-to-end training. Finally, we give additional details about our experiments on single-view reconstruction and drag minimization.

\subsection{Non-differentiability of Marching Cubes}

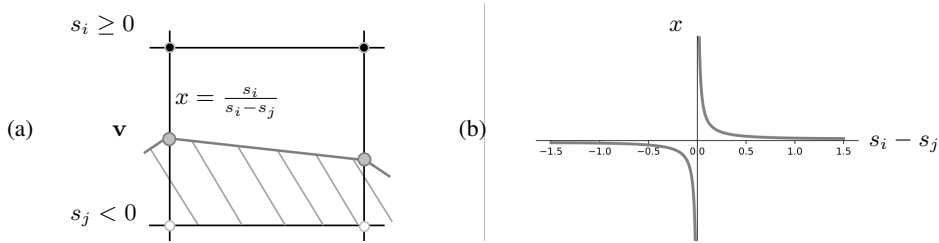
\begin{figure}[h]
\vspace{-6pt}
\hspace{6pt}
            \begin{center}
			\begin{overpic}[clip, trim=2.0cm 10cm 10cm 4cm,width=1.0\textwidth]{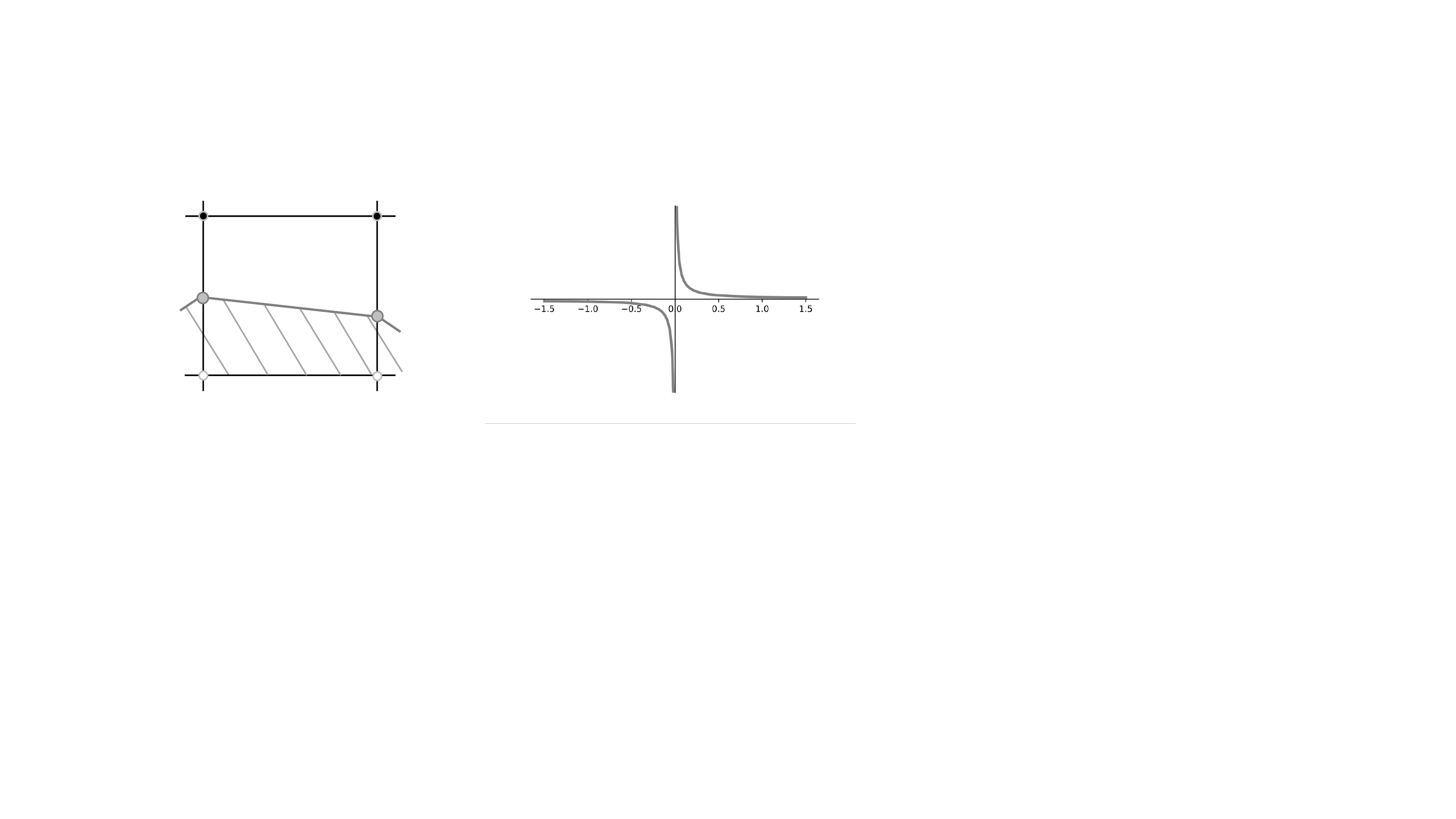}
			\put(3,20){\small{$ s _i \geq 0 $}}
			\put(3, 2){\small{$ s _j < 0 $}}
			\put(7,10){\small{$ \mathbf v $}}
			\put(13,13){\small{$ x = \frac{s _i}{s _i - s _j } $}}

			\put(-3,10){\small{(a)}}
			\put(40,10){\small{(b)}}
			\put(79,9){\small{$s_i - s_j$}}
			\put(60,20){\small{$x$}}

			\end{overpic}

			\end{center}

	\caption{\textbf{Marching cubes differentiation.} (a) Marching Cubes determines the relative position $x$ of a vertex $\bv$ along an edge via linear interpolation. This does not allow for effective back-propagation when topology changes because of a singularity when $s _i=s _j$. (b) We plot $x$, relative vertex position along an edge. Note the infinite discontinuity for $s _i=s _j$.}
		\label{fig:mc_supp}
\end{figure}

The Marching Cubes (MC) algorithm \cite{Lorensen87} extracts the zero level set of an implicit field and represents it \textit{explicitly} as a set of triangles. As discussed in the related work section, it comprises the following steps: (1) sampling the implicit field on a discrete 3D grid, (2) detecting zero-crossing of the field along grid edges, (3) assembling surface topology (i.e. the number of triangles within each cell and how they are connected) using a lookup table and (4) estimating the vertex location of each triangle by performing linear interpolation on the sampled implicit field. These steps can be understood as topology estimation followed by the determination of surface geometry.

More formally, let $S = \{ s_i \} \in \mathbb R ^ {N \times N \times N}$ denote an implicit field sampled over a discrete Euclidean grid $G_{3D} \in \mathbb R ^ {N \times N \times N \times 3}$, where $N$ denotes the resolution along each dimension. Within each voxel, surface topology is determined based on the sign of $s_{i}$ at its 8 corners. This results in $2^8 = 256$ possible surface topologies within each voxel. Once topology has been assembled, vertices are created in case the implicit field changes sign along one of the edges of the voxel.

Specifically, the vertex location $\mathbf v$ is determined using linear interpolation. Let $x \in [0,1]$ denote the vertex relative location along an edge $(\mathbf G_i, \mathbf G_j)$, where $\mathbf G_i$ and $\mathbf G_j$ are grid corners such that $s_j < 0$ and $s_i \geq 0$. This implies that if $x = 0$ then $\mathbf v = \mathbf G_i$ and conversely if  $x = 1$ then $\mathbf v = \mathbf G_j$. In the MC algorithm, $x$ is is determined as the zero crossing of the interpolant of $s_i$, $s_j$, that is, 
\begin{align}
    x = \frac{s_i}{s_i - s_j} \; .
\end{align}
Fig. \ref{fig:mc_supp}(a) depicts this process. The vertex location is then taken to be
\begin{align}
    \mathbf v = \mathbf G_i + x (\mathbf G_j - \mathbf G_i).
\end{align}
Unfortunately, this function is discontinuous for $s_i = s_j$, as illustrated in Fig \ref{fig:mc_supp}(b). Because of this, we cannot swap the signs of $s_i,s_j$ through backpropagation. This prevents topology changes while differentiating, as discussed in \cite{Liao18a}.

\subsection{Proof of Differentiable Iso-Surface Result}

\begin{figure}[t]
\vspace{-6pt}
            \begin{center}
			\begin{overpic}[clip, trim=2.0cm 10cm 10cm 2cm,width=0.7\textwidth]{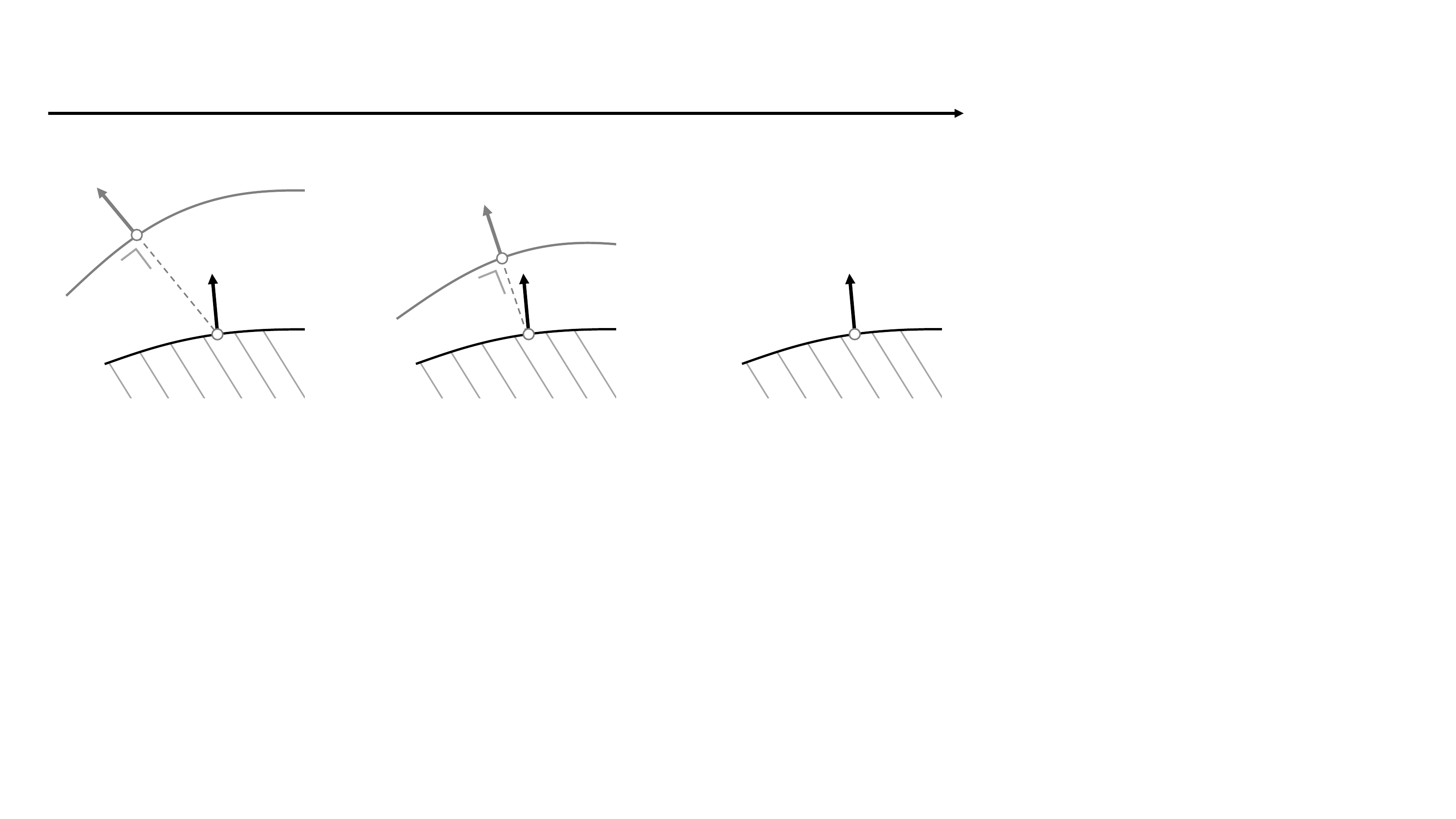}
		
			\put(-10,1){\small{$\{s = 0\} $}}
			\put(-20,10){\small{$\{s +  \Delta s = 0\} $}}
			\put(14.5,8){$\mathbf v$}
			\put(14,14){$\mathbf n (\mathbf v)$}
			\put(5,19){$\mathbf v'$}
			\put(-1,24){$\mathbf n (\mathbf v')$}
			
			\put(48,8){$\mathbf v$}
			\put(48,12){$\mathbf n (\mathbf v)$}
			\put(45,16){$\mathbf v'$}
			\put(40,21){$\mathbf n (\mathbf v')$}
			
			\put(83,8){$\mathbf v$}
			\put(77,15){$\mathbf n (\mathbf v) = - \frac{\partial \mathbf v}{\partial s} $}
			
			\put(45,31){$\Delta s \rightarrow 0$}

			\end{overpic} 
			\end{center}
	\vspace{-3mm}
	\caption{\textbf{Iso-surface differentiation.} We adopt a \textit{continuous} model in terms of how small perturbations of a signed distance function locally impact surface geometry. Here, we depict the geometric relation between local surface change $\Delta \mathbf v = \mathbf v' - \mathbf v$ and a signed distance perturbation  $\Delta s < 0$, which we exploit to compute $\frac{\partial \mathbf v}{\partial s}$ in the formal derivation below.}
		\label{fig:mc_proof}
\end{figure}
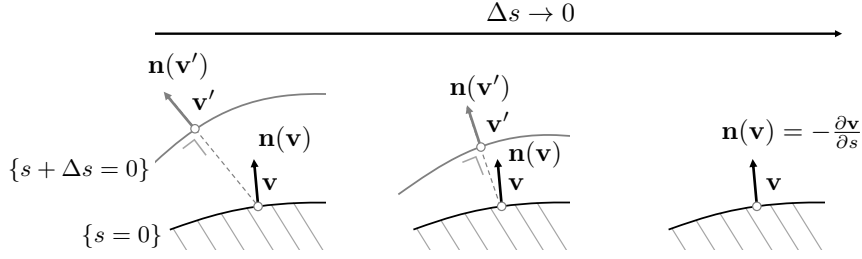

Here we formally prove Theorem 1 from the main manuscript.
\begin{theorem}
	\label{shape_derivative}
	Let us consider a signed distance function $s$ and a perturbation function $\Delta s$ such that $s+\Delta s$ is still a signed distance function. Given such $\Delta s$, we define the associated local surface change $\Delta \mathbf v = \mathbf v' - \mathbf v$ as the displacement between  $ \mathbf v'$, the closest point to  surface sample $\mathbf v $ on the perturbed surface $S' = \{ \bq \in \mathbb{R}^3 | \;  s + \Delta s (\bq) =0 \}$, and the original surface sample $\mathbf v$.
	It then holds that
	\begin{align}
	\frac{\partial \mathbf v}{\partial s } (\mathbf v) = - \mathbf n (\mathbf v) = -\nabla s (\mathbf v) \; , 
	\end{align}
	where $\mathbf n$ denotes the surface normals.
\end{theorem}
	\begin{proof}

		Recalling the definition of signed distance field, from elementary geometry we have that
    		\begin{align}
    		\Delta \mathbf v = \mathbf v' - \mathbf v =  \mathbf n(\bv ')  d(\bv , S')   = \mathbf n(\bv ')  \left(s(\mathbf \bv) - \Delta s (\mathbf \bv)\right) = -\mathbf n(\bv ')  \Delta s (\mathbf \bv).
    		\end{align}
		
		Now, observing that $ \lim _{\Delta s \rightarrow 0} \bv '  = \bv $, we have
		\begin{align}
		\frac{\partial \mathbf v}{\partial s } (\bv)=  \lim _{\Delta s \rightarrow 0} \frac{\Delta \bv }{\Delta s } = \lim _{\Delta s \rightarrow 0} -\mathbf n(\mathbf \bv') = -\mathbf n(\mathbf \bv).
		\end{align}
	Finally, recalling that for a signed distance field $\mathbf n(\bv) = \nabla s (\bv )$, follows our claim.
	\end{proof}
Fig. \ref{fig:mc_proof} illustrates this proof.

\subsection{Accelerating Iso-Surface Extraction}

Recall that our approach to iso-surface differentiation method is independent from the technique used to extract surface samples, meaning that \textit{any} non-differentiable iso-surface extraction method could be used to obtain an explicit surface from the underlying deep implicit field.

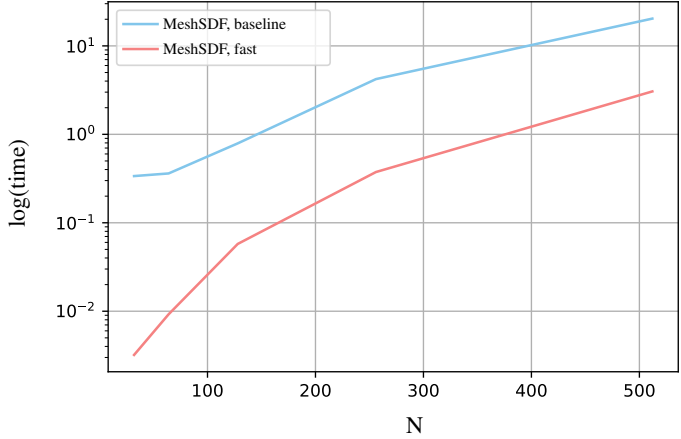
\begin{figure}[t]
            \begin{center}
			\begin{overpic}[clip, trim=0.0cm 0cm 0cm 0cm,width=0.7\textwidth]{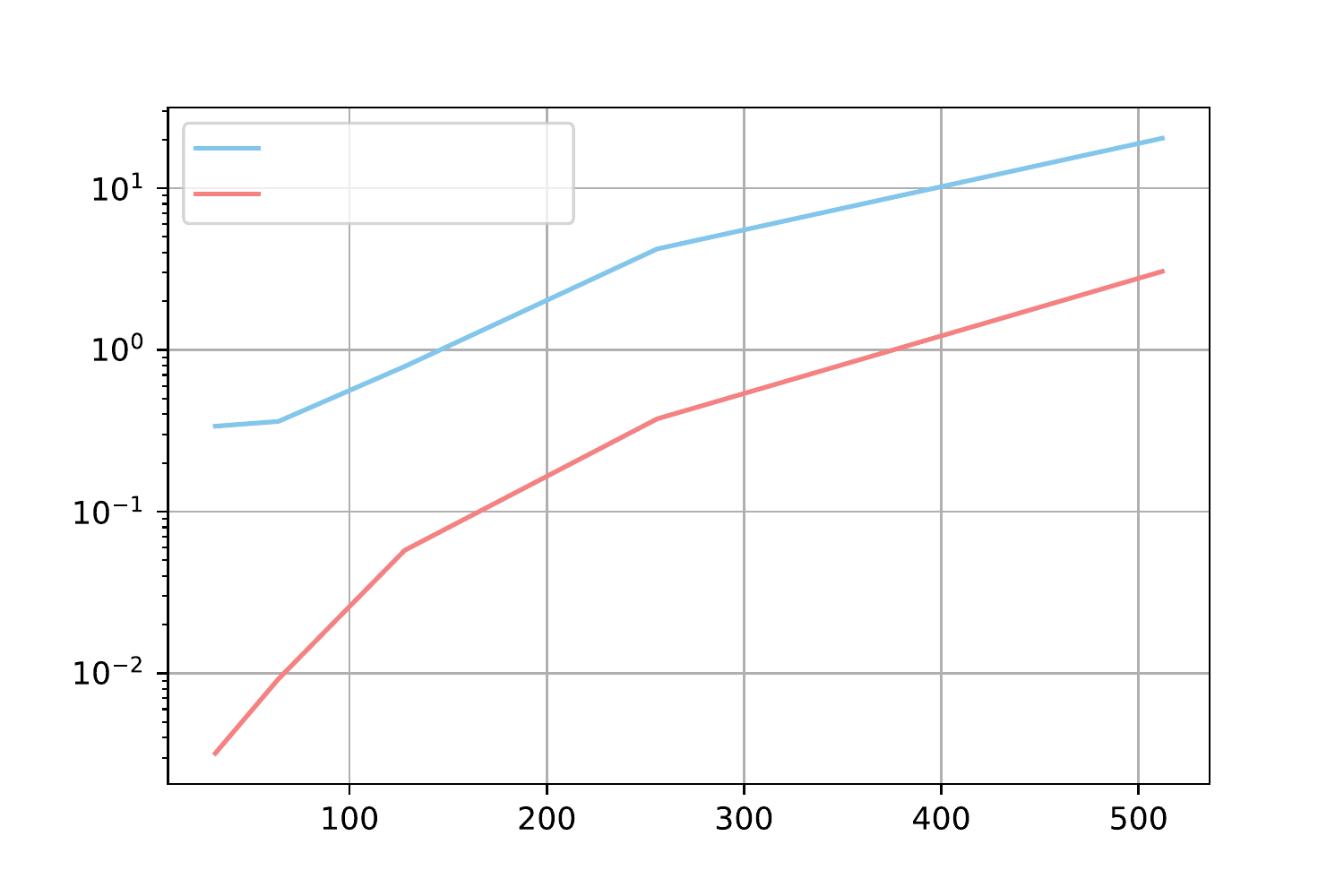}
			\put(20, 55){\tiny{MeshSDF, baseline}}
			\put(20, 51.5){\tiny{MeshSDF, fast}}

			\put(53, 0){\small{N}}
			\put(-1, 28){\rotatebox{90}{\small{log(time)}}}
			\end{overpic} 
			\end{center}
	\caption{\textbf{Accellerated Iso-Surface extraction.} When working in an iterative optimization setting, we can exploit the fact that $f_\theta$, the implicit field underlying our surface mesh representation, will change only little between iterations to evaluate it \textit{only} where we will expect it to change and consequently accelerate iso-surface extraction.}
	\label{fig:implementation}
\end{figure}

In practice, when operating in an iterative optimization settings such as those considered in the main manuscript, we exploit the fact that the deep implicit field $f_{\theta}$ is expected not to change drastically from one iteration to another, and re-evaluate it \textit{only} where we can expect new zero-crossings to appear. In this setting, we evaluate $f_{\theta}$ only at grid corners where $| f_{\theta} |$ was smaller than a given threshold at the previous iteration. This reduces the computational complexity of field-sampling from $O(N^3)$ to $O(N^2)$ in terms of the grid size $N$, which brings noticeable speed ups, as illustrated in the benchmark of Fig. \ref{fig:implementation}.

\subsection{Comparison to Deep Marching Cubes}

Deep Marching Cubes (DMC) \cite{Liao18a} is designed to convert point clouds into a surface mesh probability distribution. It can handle topological changes but is limited to low resolution surfaces for the reasons discussed in related work. In the visualization below, we compare our approach to DMC.  We fit both representations to a toy dataset consisting of two shapes: a genus-0 cow, and a genus-1 rubber duck. We use a latent space of size 2. Our metric is Chamfer $l_2$ distance evaluated on 5000 samples for unit sphere normalized shapes and shown at the bottom of the figure. As reported in the original paper, we found DMC to be unable to handle grids larger than $32^3$ because it has to keep track of all possible mesh topologies defined within the grid. By contrast, deep implicit fields are not limited in resolution and can better capture high frequency details.
\vspace{3pt}
\begin{figure}[h!]
\vspace{-2mm}
            \begin{center}
			\begin{overpic}[clip, trim=0.0cm 14cm 3cm 1cm,width=\textwidth]{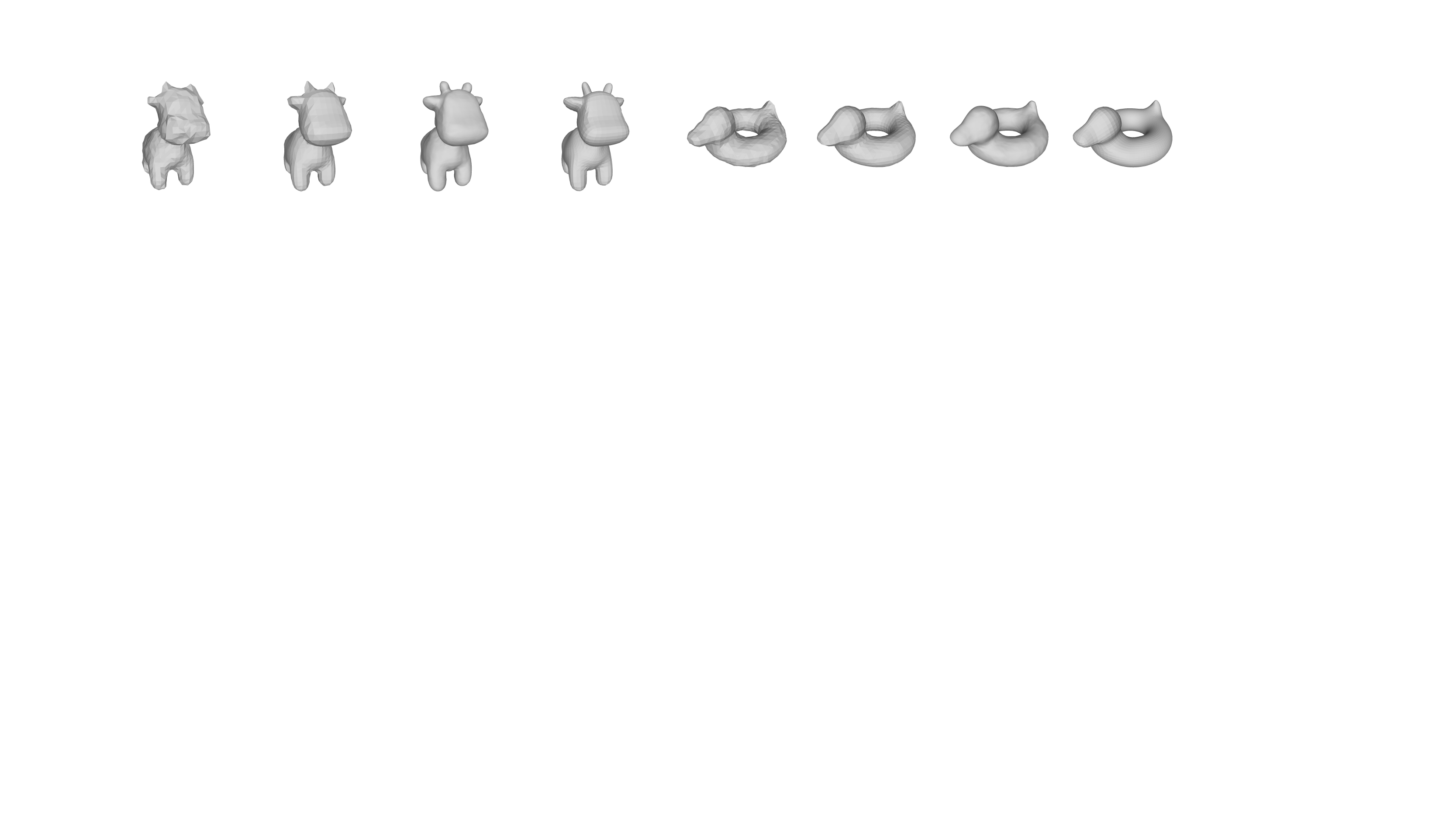}

		    \put(-2,0){\small{CD-$l_2 \cdot 10^2 \downarrow$ }}
		    
			\put(9,12){\small{DMC@$32^3$}}
			\put(11,0){\small{$1.87$}}
			\put(20,12){\small{Ours@$32^3$}}
			\put(22,0){\small{$1.84$}}
			\put(30,12){\small{Ours@256$^3$}}
			\put(32,0){{\small{{$\mathbf{1.80}$}}}}
			\put(42,12.5){\small{ground}}
			\put(43,10.7){\small{truth}}
			
			\put(51,12){\small{DMC@$32^3$}}
			\put(54,0){\small{$1.98$}}
			\put(61,12){\small{Ours@$32^3$}}
			\put(64,0){\small{$1.94$}}
			\put(71,12){\small{Ours@256$^3$}}
			\put(74,0){\small{{$\mathbf{1.90}$}}}
			\put(82,12.5){\small{ground}}
			\put(83,10.7){\small{truth}}
		
			\end{overpic} 
			\end{center}
 	\vspace{-4mm}
		\label{fig:reb}
\end{figure} 

\subsection{End-to-End Training}

Here, we demonstrate how our differentiable iso-surface extraction scheme can be used also to backpropagate gradient to the weights of MeshSDF, thus enabling end-to-end training. Specifically, let us consider a metric measuring the distance between two surfaces, such as the Chamfer $l_2$ distance
\begin{align}
\label{eq:cham}
\mathcal L_{\text{chamfer}} = \sum_{\mathbf p \in P} \min_{\mathbf q \in Q} \| \mathbf p - \mathbf q \| _2 ^ 2 + \sum_{\mathbf q \in Q} \min_{\mathbf p \in P} \| \mathbf p - \mathbf q \| _2 ^ 2 \; , 
\end{align}
where $P$ and $Q$ denote surface samples.

We exploit our differentiability result to compute 
\begin{align}
\frac{\partial \mathcal{L}_{\text{chamfer}}}{\partial \theta} &= \sum_{{\bf v} \in V}\frac{\partial \mathcal{L}_{\text{chamfer}}}{\partial {\bf v}}  \frac{\partial {\bf v}}{\partial f_{\theta}}\frac{\partial f_{\theta}}{\partial \theta} 
\\
&= \sum_{{\bf v} \in V} - \frac{\partial \mathcal{L}_{\text{chamfer}}}{\partial {\bf v}}  \nabla f_{\theta} \frac{\partial f_{\theta}}{\partial \theta} 
\; .
\label{eq:chainRuleSupp}
\end{align}
That is, we can train \textit{MeshSDF} so that to minimize directly our metric of interest. 

We evaluate the impact of doing so in Tab. \ref{table:end2end}, where we fine-tune \textit{DeepSDF} models trained minimizing the loss function $\mathcal L _{\text{sdf}}$ of the main manuscript by further minimizing $\mathcal{L}_{\text{chamfer}}$. We refer to this variant as  \textit{MeshSDF}. Unsurprisingly, fine-tuning pre-trained models by minimizing the metric of interest allows us to obtain a boost in performance. In future work, we plan to pursue the following directions within end-to-end training: increasing the level of detail in the generated surfaces by exploiting Generative Adversarial Networks operating on surface mesh data \cite{cheng2019meshgan}, and train Single View Reconstruction architectures in a semi-supervised setting, that is by using \textit{only} differentiable rasterization/rendering to supervise training.

\begin{table*}[ht]
	\centering
	\small
	\caption{\textbf{End-to-end training.} We exploit end-to-end-differentiability to fine-tune pre-trained \textit{DeepSDF} networks so that to that to minimize directly our metric of interest, Chamfer distance.}
	\label{table:end2end}
	\begin{tabularx}{\linewidth}{c| c c | c c }
		\toprule
		\myfootnotesize Category & \myfootnotesize DeepSDF(train) & \myfootnotesize \myfootnotesize MeshSDF(train) &\myfootnotesize \myfootnotesize DeepSDF(test) & \myfootnotesize \myfootnotesize MeshSDF(test)\\
		\midrule
		Cars   &  0.00071 & \textbf{0.00064} ($\downarrow 9 \%$)   & 0.00084  & \textbf{0.00067} ($\downarrow 20 \%$)\\
		Chairs &  0.00145 & \textbf{0.00133} ($\downarrow 8 \%$)  & 0.00407  & \textbf{0.00259} ($\downarrow 36 \%$) \\
		\bottomrule
	\end{tabularx}
\end{table*}

\subsection{Single View Reconstruction}

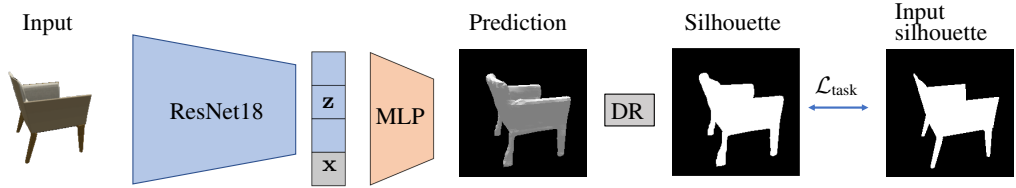
\begin{figure}[t]
        \vspace{-8pt}
		\begin{center}
			\begin{overpic}[clip, trim=0.0cm 7cm 0 4.5cm,width=1.0\textwidth]{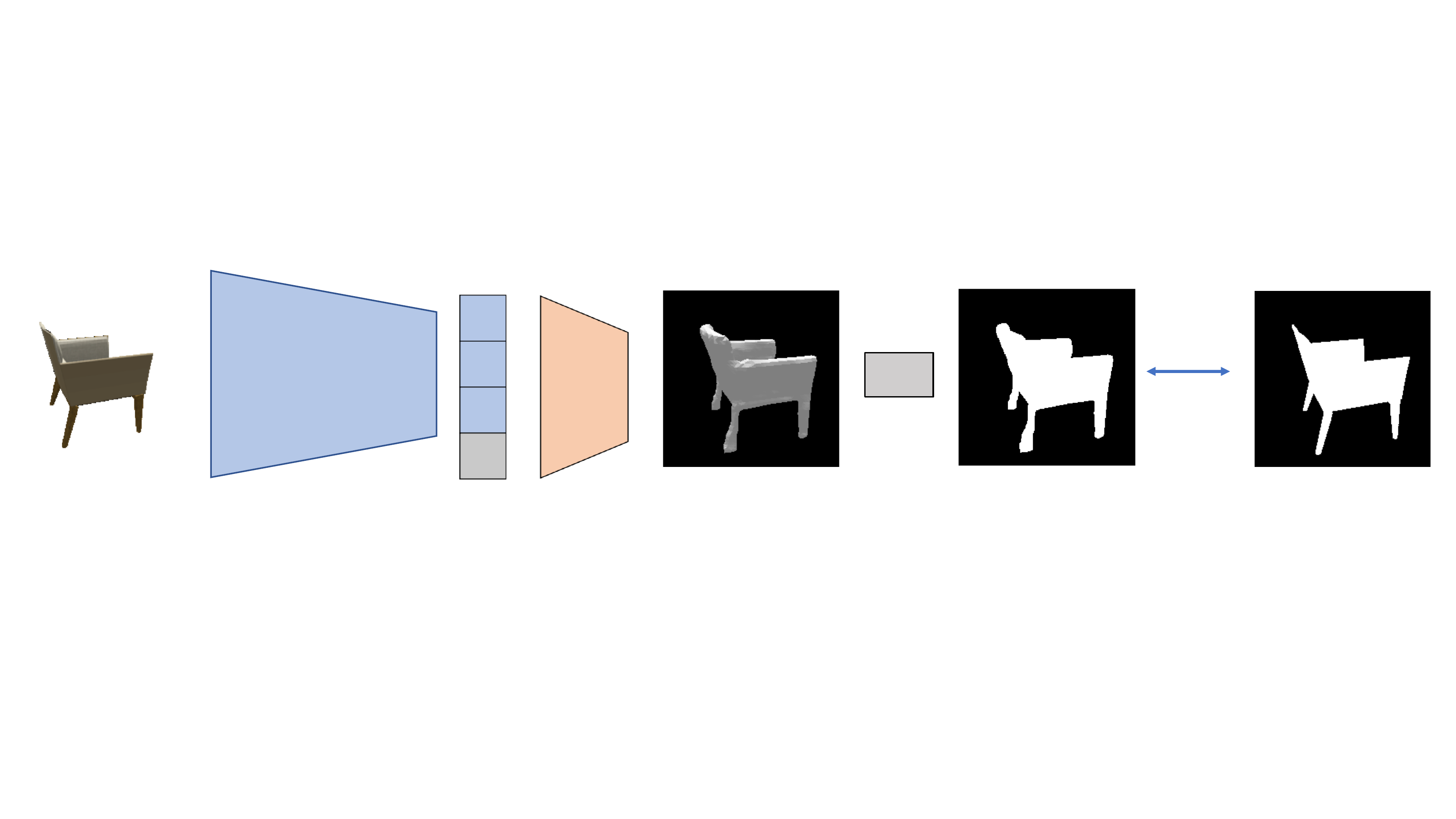}
			\put(4,17.5){\small{Input }}
			\put(18,9){\small{ResNet18}}
			\put(37.6,8.5){\small{MLP}}
			
			\put(32.5,10){\small{$\mathbf z$}}
			\put(32.5,4){\small{$\mathbf x$}}
			
			\put(46.5,17.5){\small{Prediction}}

			\put(60,9){\small{DR}}
			
			\put(67,17.5){\small{Silhouette}}
			
			\put(79.4,11.5){\small{$\mathcal{L}_{\text{task}}$}}
			
			\put(87,18.5){\small{Input}}
			\put(87,16.5){\small{silhouette}}
			
			\end{overpic}
		\end{center}
		\vspace{-9pt}
		\caption{\textbf{Shilouette-driven refinment.} At inference time, given an input image, we exploit the differentiability of \textit{MeshSDF} to refine the predicted surface so that to match input silhouette in image space through Differentiable Rasterization \cite{kato2018neural}.}
\label{fig:svr_archi}
\vspace{-6pt}
\end{figure}

We first provide additional details on the Single View Reconstruction pipeline presented in the main manuscript. Then, for each experimental evaluation of the main paper, we first introduce metrics in details, and then provide additional qualitative results. To foster reproducibility, we will make our entire code-base publicly available.

\paragraph{Architecture.} 

Fig \ref{fig:svr_archi} depicts our full pipeline. As in earlier work~\cite{Mescheder19,Chen19c}, we condition our deep implicit field architecture on the input images via a residual image encoder~\cite{He_2016}, which maps input images to latent code vectors $\bz$.  Specifically, our encoder consists of a ResNet18 network, where we replace batch-normalization layers with instance normalization ones \cite{ulyanov2016instance} so that to make harder for the network to use color cues to guide reconstruction. 
These latent codes are then used to condition the signed distance function Multi-Layer Perceptron (MLP) architecture of the main manuscript, consisting of 8 Perceptrons as well as residual connections, similarly to \cite{Park20a}. We train this architecture, which we dub \emph{MeshSDF} (raw), by minimizing $\mathcal L_{\text{sdf}}$ (Eq.1 on the main manuscript) wrt. $\theta$ on a training set of image-surface pairs. 

At inference time, we exploit end-to-end differentiability to refine predictions as depicted in Fig \ref{fig:svr_archi}. 
That is, given the camera pose associated to the image and the current value of $\bz$, we project vertices and facets into a binary silhouette in image space through a differentiable rasterization function $\text{DR}_{\text{silhouette}}$~\cite{kato2018neural}. Ideally, the projection matches the observed object silhouette $\cal S$ in the image, which is why we define our objective function as
\begin{equation}
\mathcal{L}_{\text{task}} = \| \text{DR}_{\text{silhouette}}(\mathcal M(\bz))- \mathcal S\|_1 \:, 
\end{equation}
which we minimize with respect to $\bz$. In practice, we run 400 gradient descent iterations using Adam~\cite{kingma2014adam} and keep the $\bz$ with the smallest $\mathcal{L}_{\text{task}}$ as our final code vector. 

\paragraph{Evaluation on ShapeNet.} 

We used standard train/test splits along with the renderings provided in~\cite{xu2019disn} for all the comparisons we report. We evaluate different approaches based on the following SVR metrics:
\begin{itemize}
  \item \textbf{Chamfer $l_2$ pseudo-distance:} Common evaluation metric for measuring the distance between two uniformly sampled clouds of points $P,Q$, defined as 
  \begin{align}
\text{CD-$l_2$}(P,Q) = \sum_{\mathbf p \in P} \min_{\mathbf q \in Q} \| \mathbf p - \mathbf q \| _2 ^ 2 + \sum_{\mathbf q \in Q} \min_{\mathbf p \in P} \| \mathbf p - \mathbf q \| _2 ^ 2.
\end{align}
We evaluate this metric by sampling $2048$ points from reconstructed and target shape, which are re-scaled to fit into a unit-radius sphere.

  \item \textbf{Earth Mover distance:} This metric measures the distance between two point clouds by solving an assignment problem
  \begin{align}
      {\text{EMD}}(P,Q) = \min_{\Phi : \, P \rightarrow Q} \sum_{\mathbf p \in P} \| \mathbf p - \Phi( \mathbf p) \| _2, 
  \end{align}
  where, for all but a zero-measure subset of point set pairs, the optimal bijection $ \Phi $
is unique and invariant under infinitesimal movement of the
points. In practice, the exact computation of EMD is too expensive and we implement the $(1 + \varepsilon )$ approximation scheme of \cite{Fan17a}.
We evaluate this metric by sampling $2048$ points from reconstructed and target shape, which are re-scaled to fit into a unit-radius sphere.

  \item \textbf{Intersection over Union:} Since all information about an object’s shape is situated on its surface, and to allow comparison to methods that do not produce watertight surfaces (such as \cite{groueix2018papier}) we propose to evaluate object similarity by measuring surface-to-surface IoU. In practice, denoting as $\mathcal V$ the function mapping a cloud of points to a binary voxel grid, this metric reads
  \begin{align}
      {\text{IoU}}(P,Q) = \frac{\text{intersection} (\mathcal V (P), \mathcal V (Q) )}{\text{union} (\mathcal V (P), \mathcal V (Q) )} 
  \end{align}
  We evaluate this metric by sampling $5000$ points and setting up the voxel grid divide the object bounding box at resolution $50\times50\times50$.
  
  \item \textbf{F-score:} The F-Score has been recently proposed \cite{Tatarchenko19} for evaluating SVR algorithms. It explicitly evaluates the distance between object surfaces and is defined as the harmonic mean between precision and recall at a given distance threshold $d$. We refer the reader to \cite{Tatarchenko19} for more details about this metric. We evaluate this metric by sampling $10000$ points from reconstructed and target shape and set $5\%$ of the object bounding box length as distance threshold.
\end{itemize}

In Table \ref{tab:shapenet_supp}, we further compare our method to state-of-the-art single view reconstruction algorithms in terms of F-score. Similarly to what reported in the main manuscript for CD, EMD and IoU, performing imaged-based refinement allows us to outperforms all other state-of-the-art approaches also in terms of this metric.  
\begin{table*}[h]
    \caption{\textbf{Single view reconstruction results on ShapeNet Core.} Exploiting end-to-end differentiability to perform image-based refinement allows us to outperform all prior methods also in terms of F-Score.}
	\label{tab:shapenet_supp}
	\begin{center}
		\scalebox{0.65}{
			\begin{tabular}{c|c|ccccccccccccc|c}
				\Xhline{2\arrayrulewidth}
				Metric & Method & plane & bench & cabinet & car & chair & display & lamp & speaker & rifle & sofa & table & phone & boat & mean \\
				\Xhline{2\arrayrulewidth}
				\multirow{6}{*}{F-Score$\%$ $\uparrow$}&AtlasNet &91 & 86 &74 &94 &91 &84 &81 &80 &96 &91 &91 &90 &90 & 89 \\
                &Pixel2Mesh&88 & 95 & \textbf{94} & 97 & 94 & 92 & 89 & 89 & 95 & \textbf{96} & 93 & 97 & 94 & 93 \\
                &Mesh R-CNN& 87&91 &90 &95 & 90 & 89 & 83 & 85 & 93 & 92 & 90 & 95 & 91 & 90 \\
                &DISN&94&94&89&96&90&92&78&85&96&\textbf{96}&87&96&93&91 \\
				\cline{2-16}
				&MeshSDF(raw)&92&95&92&98&94&91&85&86&96&94&91&95&93&91\\
				&MeshSDF&\textbf{96}&\textbf{97}&\textbf{94}&\textbf{98}&\textbf{97}&\textbf{95}&\textbf{91}&\textbf{91}&\textbf{98}&\textbf{96}&\textbf{94}&\textbf{98}&\textbf{95}&\textbf{95}\\
				\Xhline{2\arrayrulewidth}
		\end{tabular}}
	\end{center}
\end{table*}

\paragraph{Evaluation on Pix3D.} 

We followed closely the evaluation pipeline proposed together with this dataset \cite{Sun2018Pix3DDA}. 
That is, we focus on the chair category, and exclude from the evaluation all images where the object we want to reconstruct is truncated or occluded, resulting in $2894$ test images. 
We then use ground truth bounding boxes to crop the image to a window centered around the object.
To evaluate fairly reconstruction performance, we segment the background off for all methods presented in Table 2 of the main paper but for \cite{Sun2018Pix3DDA}, that achieves state-of-the-art performance in joint segmentation and reconstruction on this benchmark. We do so to give a sense of the impact of assuming to have accurate segmentation information on reconstruction quality.
Finally, following the evaluation pipeline designed in \cite{Sun2018Pix3DDA}, we only have access to ShapeNet synthetic data to train our models, that is we don't have access to any Pix3D image at training time. The main challenge of this benchmark is therefore to design an architecture that is robust to the change of domain.
Finally, we use evaluation metrics as originally proposed in \cite{Sun2018Pix3DDA}:

\begin{itemize}
  \item \textbf{Chamfer $\sqrt{l_2}$ pseudo-distance:}
  \begin{align}
\text{CD-$\sqrt{l_2}$}(P,Q) = \sum_{\mathbf p \in P} \min_{\mathbf q \in Q} \| \mathbf p - \mathbf q \| _2  + \sum_{\mathbf q \in Q} \min_{\mathbf p \in P} \| \mathbf p - \mathbf q \| _2,
\end{align}
where $P$ and $Q$ are clouds of points.
We evaluate this metric by sampling $1024$ points from reconstructed and target shape, which are re-scaled to fit into a $[-0.5, 0.5]^3$ bounding box. \footnote{In the main manuscript we have dubbed this metric as Chamfer ${l_1}$ by mistake. We will fix when we revise the paper.}

  \item \textbf{Earth Mover distance:} We use the same metric as above, but follow the approximation scheme of \cite{Sun2018Pix3DDA} in this case.
We evaluate this metric by sampling $1024$ points from reconstructed and target shape, which are re-scaled to fit into a $[-0.5, 0.5]^3$ bounding box.

  \item \textbf{Intersection over Union:} We evaluate object similarity by measuring volume-to-volume IoU. In practice, denoting as $\mathcal F$ the function mapping a surface mesh $\mathcal M$ to a filled-in binary voxel grid, this metric reads
  \begin{align}
      {\text{IoU}_{\text{vol}}}(\mathcal P, \mathcal Q) = \frac{\text{intersection} (\mathcal F (\mathcal P), \mathcal F (\mathcal Q) )}{\text{union} (\mathcal F (\mathcal P), \mathcal F (\mathcal Q) )},
  \end{align}
  where $\mathcal P, \mathcal Q$ denote surface meshes. We evaluate this metric by setting up the voxel grid divide the surface mesh bounding box at resolution $32\times32\times32$.
  
\end{itemize}

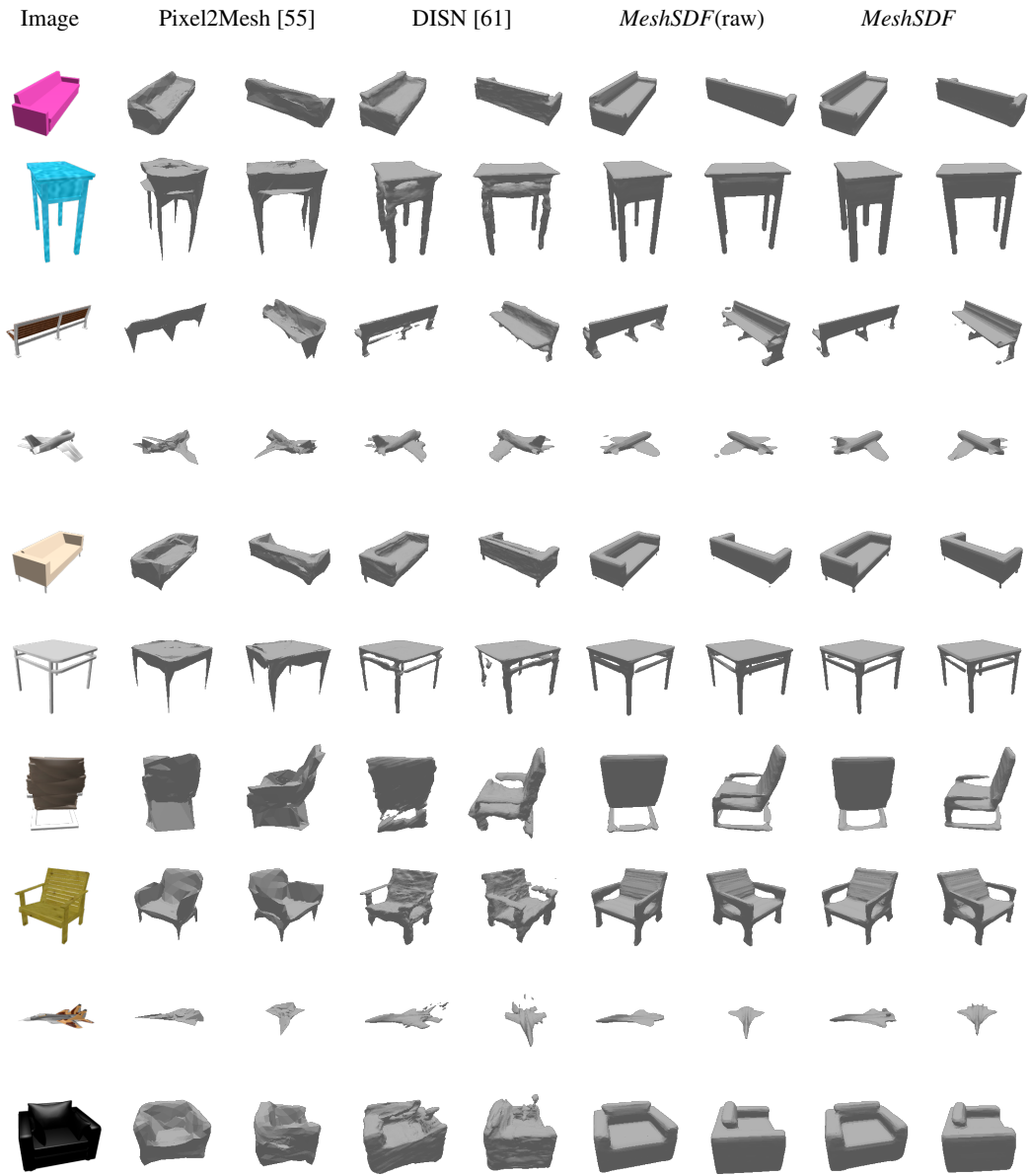
\begin{figure}[h!]
        \vspace{25pt}
		\begin{center}
			\begin{overpic}[clip, trim=0.0cm 0cm 0 0.0cm,width=1.0\textwidth]{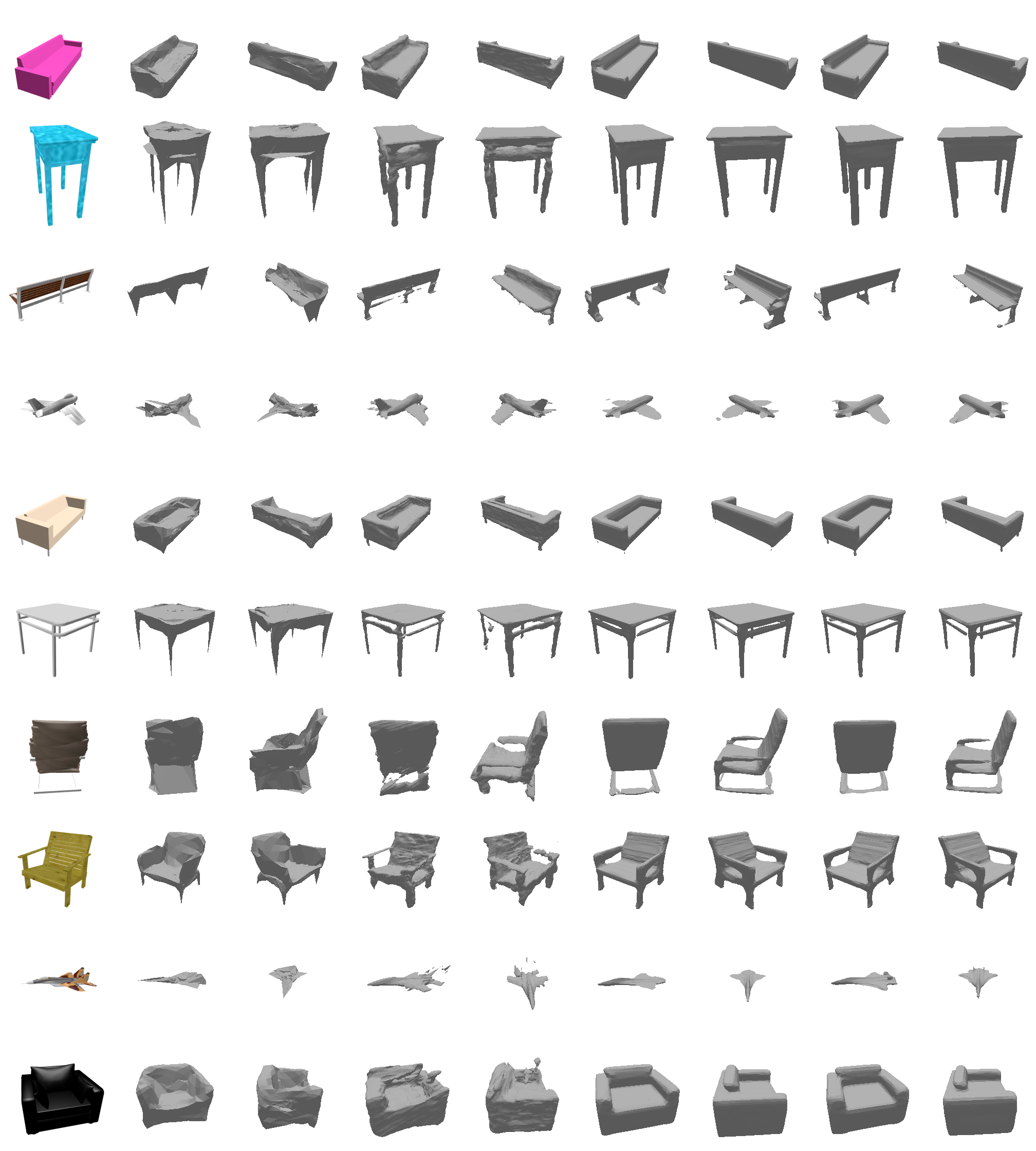}
			\put(2,101){\small{Image }}
			\put(14,101){\small{Pixel2Mesh \cite{Wang_2018}}}
		    \put(36,101){\small{DISN \cite{xu2019disn}}}
			\put(54,101){\small{\textit{MeshSDF}(raw)}}
			\put(75,101){\small{\textit{MeshSDF}}}
			\end{overpic}
		\end{center}
		\vspace{-10pt}
		\caption{\textbf{Comparative results for SVR on ShapeNet.} }
\label{fig:ShapeNet_qual}
\vspace{-6pt}
\end{figure}

\textbf{Additional qualitative results.} We provide additional qualitative comparative results on both ShapeNet and Pix3D  in Fig \ref{fig:ShapeNet_qual},\ref{fig:Pix3D_qual}. Furthermore, in Fig \ref{fig:failure} we show failure cases, which we obtain by selecting samples for which refinement does not bring any improvement. Furthermore, we refer the reader to the supplementary video for animations depicting the impact of iterative refinement on the reconstruction.

\subsection{Aerodynamic Shape Optimization}

\begin{figure}[t]
		\vspace{-3pt}
		\begin{center}
			\begin{overpic}[clip, trim=0.1cm 3.5cm 0.1cm 3cm,width= \textwidth]{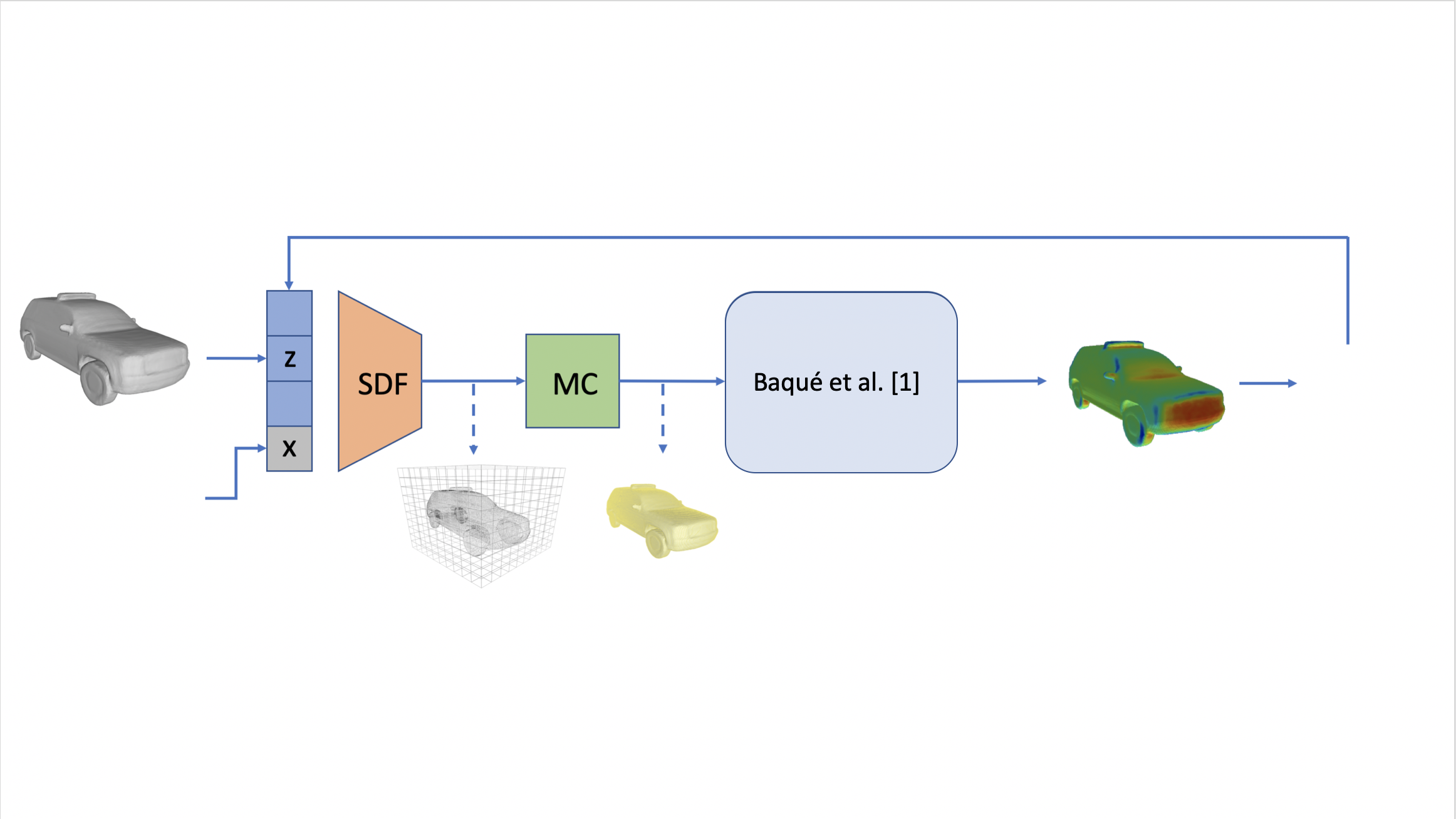}
			    \put(3,13){\small{Grid Points}}
			    
			    \put(28,4){\tiny{Predicted SDF}}
			    
			    \put(40,4){\tiny{Reconstructed Mesh}}
			    
			    \put(40,33){\small{MeshSDF Gradient (Theorem 1)}}
			    
    			\put(90,21){$\mathcal L_{\text{task}}$}
			\end{overpic}
		\end{center}
		\vspace{-6pt}
		\caption{\textbf{Aerodynamic optimization pipeline.}
		     We encode a shape we want to optimize using \textit{DeepSDF} (denoted as \textbf{SDF} block on the figure) and obtain latent code $\mathbf z$. Then we start our iterative process. First, we assemble an Euclidean grid and predict SDF values for each node of the grid. On this grid we run the Marching Cubes algorithm (\textbf{MC}) to extract a surface mesh. We then run the obtained shape through a Mesh CNN (\textbf{CFD}) to predict pressure field from which we compute drag as our objective function. Using the proposed algorithm we obtain gradients of the objective w.r.t. latent code $z$ and do an optimization step. The loop is repeated until convergence. 
		}
\label{fig:cfd_pipeline}
\end{figure}
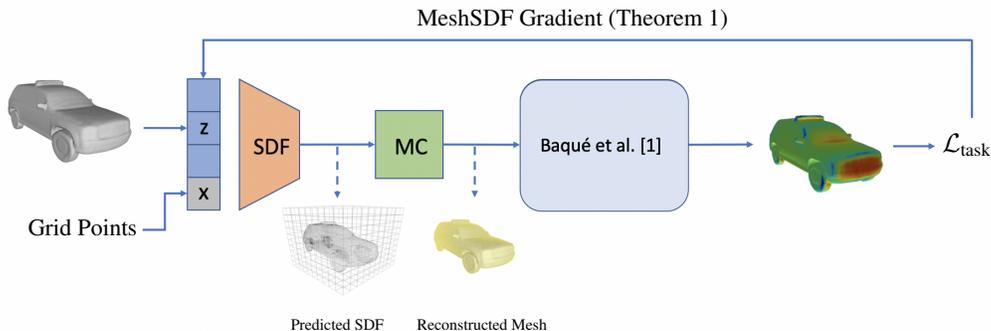

Here we provide more details on how we performed the aerodynamic optimization experiments presented in the main manuscript.  The overall pipeline for the optimisation process is depicted in Fig.~\ref{fig:cfd_pipeline}, and additional optimization results are shown in Fig.~\ref{fig:cfd_opta}.

\subsubsection{Dataset}

As described in the main manuscript, we consider the car split of the ShapeNet~\cite{chang2015shapenet} dataset for this experiment.  Since aerodynamic simulators typically require high quality surface triangulations to perform CFD simulations reliably, we (1) follow~\cite{sin2013vega} and automatically remove internal part of each mesh as well as re-triangulate surfaces and (2) manually filter out corrupted surfaces. After that, we train a DeepSDF auto-decoder on the obtained data split and, using this model, we reconstruct the whole dataset from the learned parameterization. The last step is needed so that to provide fair initial conditions for each method of the comparison in Tab. 3 of the main manuscript, that is to allow all approaches to begin optimization from identical meshes.

We obtain ground truth pressure values for each car shape with OpenFoam~\cite{jasak2007openfoam}, setting an \textit{inflow velocity} of $15$ meters per second and airflow \textit{density} equal $1.18$.
Each simulation was run for at most $5000$ time steps and took approximately 20 minutes to converge. 
Some result of the CFD simulations are depicted in the top row of Fig.~\ref{fig:cfd_prediction}.

We will make both the cleaned car split of ShapeNet and the simulated pressure values publicly available.

\subsubsection{CFD prediction}

We train a Mesh Convolutional Neural Network to regress pressure values given an input surface mesh, and then compute aerodynamic drag by integrating the regressed field. Specifically, we used the dense branch of the architecture proposed in~\cite{Baque18} and replaced Geodesic Convolutions~\cite{monti2017geometric} by Spline ones~\cite{Fey/etal/2018} for efficiency.

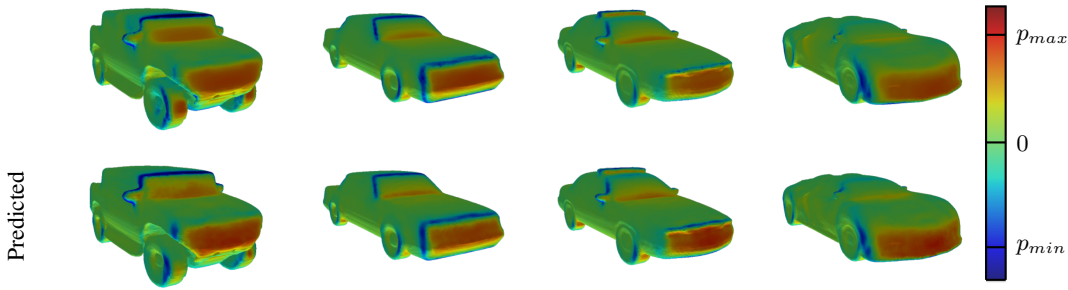
\begin{figure}[t]
		\vspace{-3pt}
		\begin{center}
			\begin{overpic}[clip, trim=0.0cm 1.5cm 0cm 5cm,width= \textwidth]{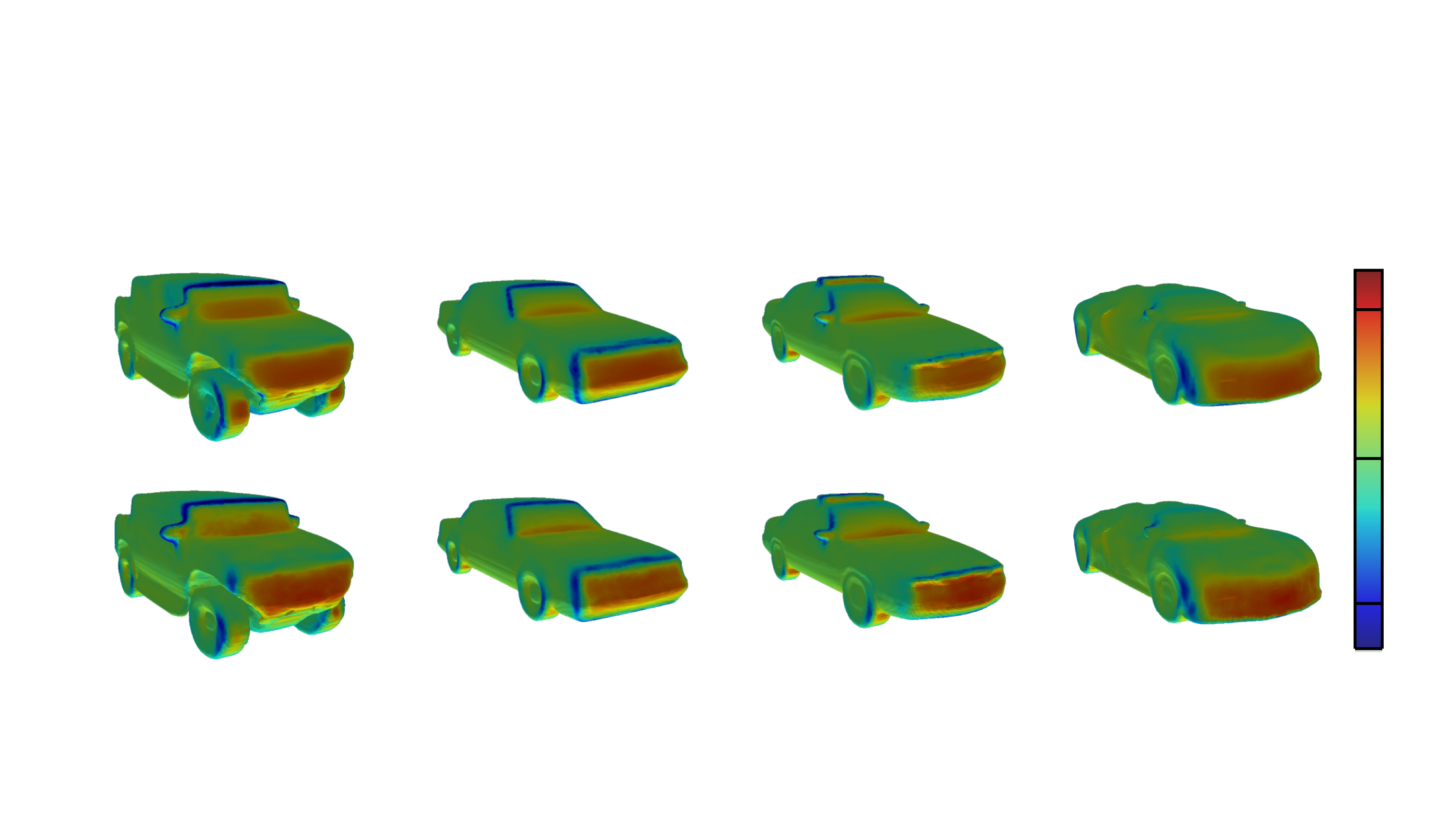}
    			\put(0, 24){}
    			\put(0, 9){\rotatebox{90}{\small{Predicted}}}
    			
    			\put(96, 10){\small{$p_{min}$}}
    			\put(96, 19.5){\small{$0$}}
    			\put(96, 30){\small{$p_{max}$}}
			\end{overpic}
		\end{center}
		\vspace{-30pt}
		\caption{\textbf{Simulated and predicted pressure fields.}
		    Pressure fields for different shapes simulated with OpenFoam (top) and predicted by the Mesh Convolutional Neural Network (bottom). 
		}
\label{fig:cfd_prediction}
\end{figure}

A comparison for the predicted and simulated pressure values may be seen in Fig.~\ref{fig:cfd_prediction}.

\subsubsection{Implementation Details}

In this section we provide the details needed to implement the baselines parameterizations presented in the main manuscript. \comment{To foster reproducibility, we will make our code publicly available.}

\begin{itemize}
    \item\textbf{Vertex-vise optimization}
    In this baseline, we optimize surface geometry by flowing gradients directly into surface mesh vertices, that is without using a low-dimensional parameterization.
 In our experiments, we have found this strategy to produce unrealistic designs akin to adversarial attacks that, although are minimizing the drag predicted by the network, result in CFD simulations that do not convergence.
    This confirms the need of using a low-dimensional parameterization to regularize optimization.

    \item\textbf{Scaling}
    We apply a function $f_{C_x, C_y, C_z} (V) = (C_x V_x, C_y V_y, C_z V_z)^T$ to each vertex of the initial shape.
    Here $C_i$ are 3 parameters describing how to scale vertex coordinates along the corresponding axis.
    As we may see from the Tab. 3 of the main manuscript, such a simple parameterization already allows to improve our metric of interest.

    \item\textbf{FreeForm}
    Freeform deformation is a very popular class of approaches in engineering optimization.
    A variant of this parameterization was introduced in~\cite{Baque18}, where it led to good design performances.
    In our experiments we are using the parameterization described in~\cite{Baque18} with only a small modification:
    to enforce the car left and right sides to be symmetrical we square sinuses in the corresponding terms.
    We also add $l_2$-norm of the parameterization vector to the loss as a regularization.

    \item\textbf{PolyCube}
    Inspired by~\cite{Umetani18} we create a grid of control points to change the mesh.
    The grid size is $8 \times 8 \times 8$ and it is aligned to have $20\%$ width padding along each axis.
    The displacement of each control point is limited to the size of each grid cell, by applying $tanh$.
    During the optimization we shift each control point depending on the gradient it has and then tri-linearly interpolate the displacement to corresponding vertices.
    Finally, we enforce the displacement field to be regular by using Gaussian Smoothing ($\sigma = 1$, kernel size $= 3$). This results in a parameterization that allows for deformations that are very similar to the one of ~\cite{Umetani18}.
    \comment{
    Finally, to make the surface smooth, on top of the linear interpolation we use Gaussian Smoothing.}

\end{itemize}

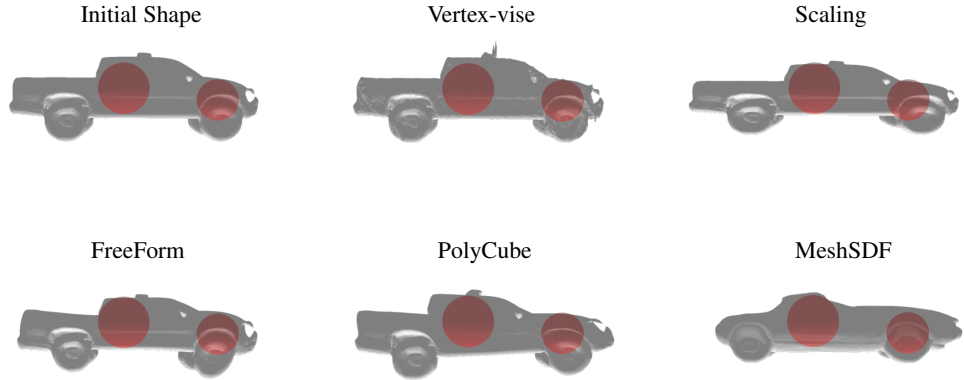
\begin{figure}[t]
		\vspace{-3pt}
		\begin{center}
			\begin{overpic}[clip, trim=4cm 3.5cm 3cm 2.5cm,width= \textwidth]{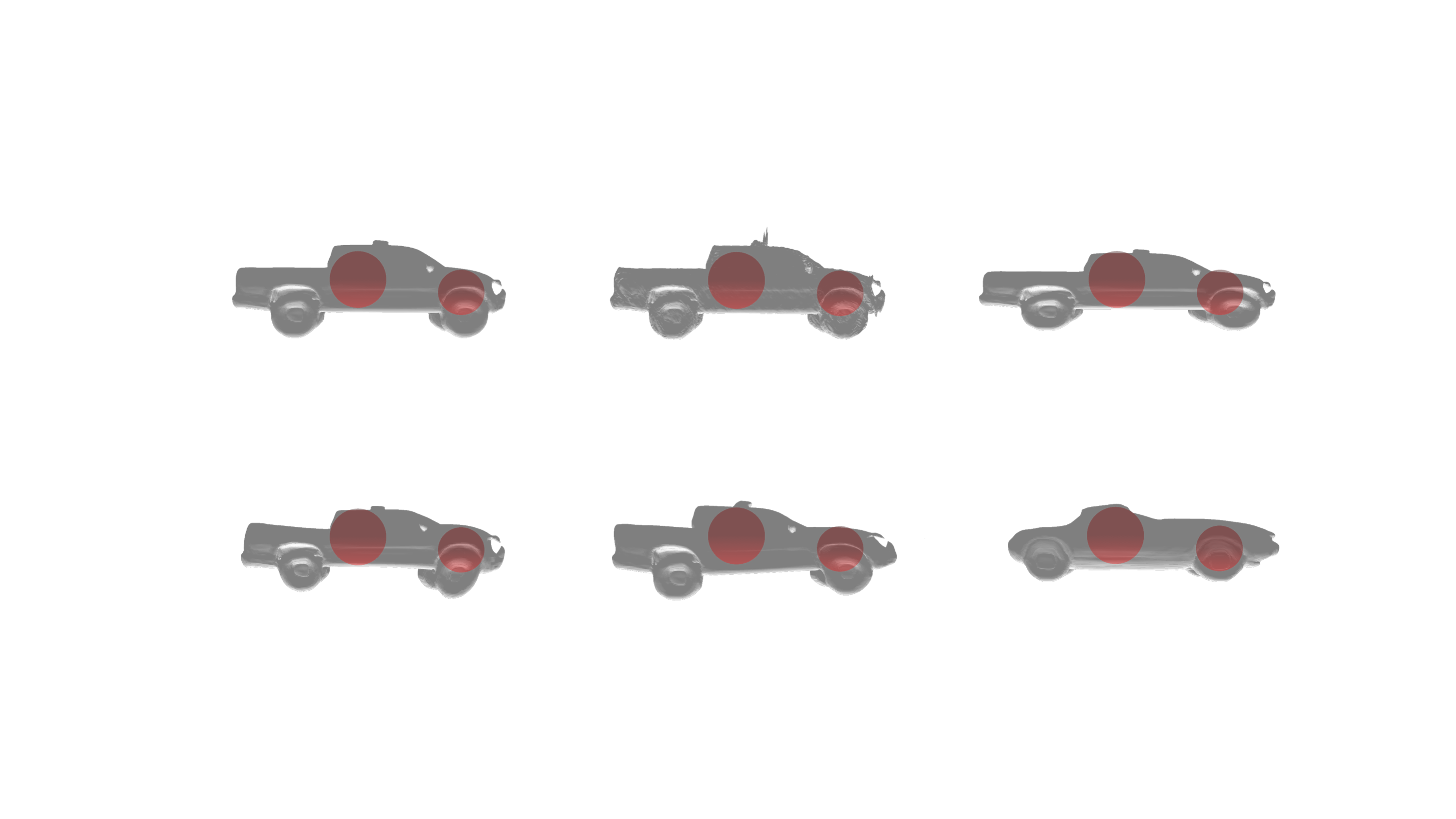}
			   \put(12,40){\small{Initial Shape}}
			   \put(45,40){\small{Vertex-vise}}
			   \put(80,40){\small{Scaling}}
			   
			   \put(13,17.5){\small{FreeForm}}
			   \put(46,17.5){\small{PolyCube}}
			   \put(80,17.5){\small{MeshSDF}}
			    
			\end{overpic}
		\end{center}
		\caption{\textbf{Soft constraints reserving space for driver and engine.}
		    The figure illustrates the constraints we put on the surfaces during the optimization process.
		    The constraints are shown for the initial shape, and then for all presented parameterizations.
		    Note, that the constraints we put are soft, and thus may be violated.
		}
\label{fig:cfd_constraints}
\end{figure}

As we describe in the main paper, to prevent the surface from collapsing to a point, we put a set of soft-constraints to reserve space for driver and engine.
The constraints are represented on the figure~\ref{fig:cfd_constraints}.

\subsubsection{Additional Regularization for MeshSDF}

In order to avoid generating unrealistic designs with MeshSDF, we introduce an additional regularization term $\mathcal L _{\text{constraint}}$ in the optimization, similarly to the regularizations introduced in the baseline parameterizations discussed above.

In our experiments, we began by using a standard penalty on $l_2$ norm of the latent code, $\mathcal L _{\text{constraint}} = \alpha ||\mathbf z||^2_2$.
However, even though it prevented most of the runs from converging to unrealistic shapes, we found converged shapes to still be coarse and noisy in some cases.

We therefore opted for a more conservative regularization strategy, reading
\begin{align}
\mathcal L _{\text{constraint}} = \alpha \sum_{\mathbf z' \in \mathcal{Z}_k} \frac{||\mathbf z - \mathbf z'||_2^2}{|\mathcal{Z}_k|},
\end{align}
where $\mathcal{Z}_k = {\mathbf z_0, \mathbf z_1, \ldots, \mathbf z_k}$ denote the $k$ closest latent vectors to $\mathbf z$ from the training set of DeepSDF.
In our experiments we set $k = 10$, $\alpha = 0.2$.
This regularization limits exploration of the latent space, but guarantees more robust and realistic optimisation outcomes.

In our aerodynamics optimization experiments, different initial shapes yield different final ones. 
We speculate that this behavior is due to the presence of local minima in the latent space of DeepSDF, even though we use the Adam optimizer~\cite{kingma2014adam} , which is known for its ability to escape some of them.
We are planning to address the problem more thoroughly in future.

\subsection{Comparison to implicit field differentiable rendering}
Recent advances in differentiable rendering ~\cite{liu2020dist} have shown that is possible to render continuous SDFs differentiably by carefully designing a differentiable version of the sphere tracing algorithm. 
By contrast, we simply use MeshSDF end-to-end differentiability to exploit an \textit{off-the-shelf} differentiable rasterizer to achieve the same result.
To highlight the advantages of doing so, we take the generative model of Section 1.4, initialize latent code so that to generate the cow, and then minimize silhouette distance with respect to the duck. In the table below we compare our approach to \cite{liu2020dist}.
Sphere tracing requires to query the network along each camera ray in a sequential fashion, resulting in longer computational time with respect to our approach, which projects surface triangles to image space and then rasterizes them in parallel. Furthermore, our approach requires less function evaluation, as we do not need to sample densely the volume around the field zero-crossing. 

\begin{table*}[h]
    \vspace{-8pt}
	\centering
	\small
	\begin{tabularx}{\linewidth}{@{}X*4{@{\hspace{2mm}}c}@{\hspace{3mm}}c@{\hspace{2mm}}c@{}}
		\toprule
		\myfootnotesize Method & \myfootnotesize $l_2$ silhouette distance $\downarrow$  & \myfootnotesize  \# network queries $\downarrow$   & \myfootnotesize run time [s] $\downarrow$ \\
		\midrule
		Liu20 [most efficient settings, $512^2$ renders]  & 0.005973 &  898k  & $1.24$ \\
		MeshSDF [isosurface at $256^3$, $512^2$ renders]
		& \textbf{0.004625}  & \textbf{266k} & \textbf{0.29} \\
		\bottomrule
	\end{tabularx}
	\vspace{-3pt}
\end{table*}


\begin{figure}[h!]
        \vspace{-8pt}
		\begin{center}
			\begin{overpic}[clip, trim=0.0cm 0cm 0 0.0cm,width=1.0\textwidth]{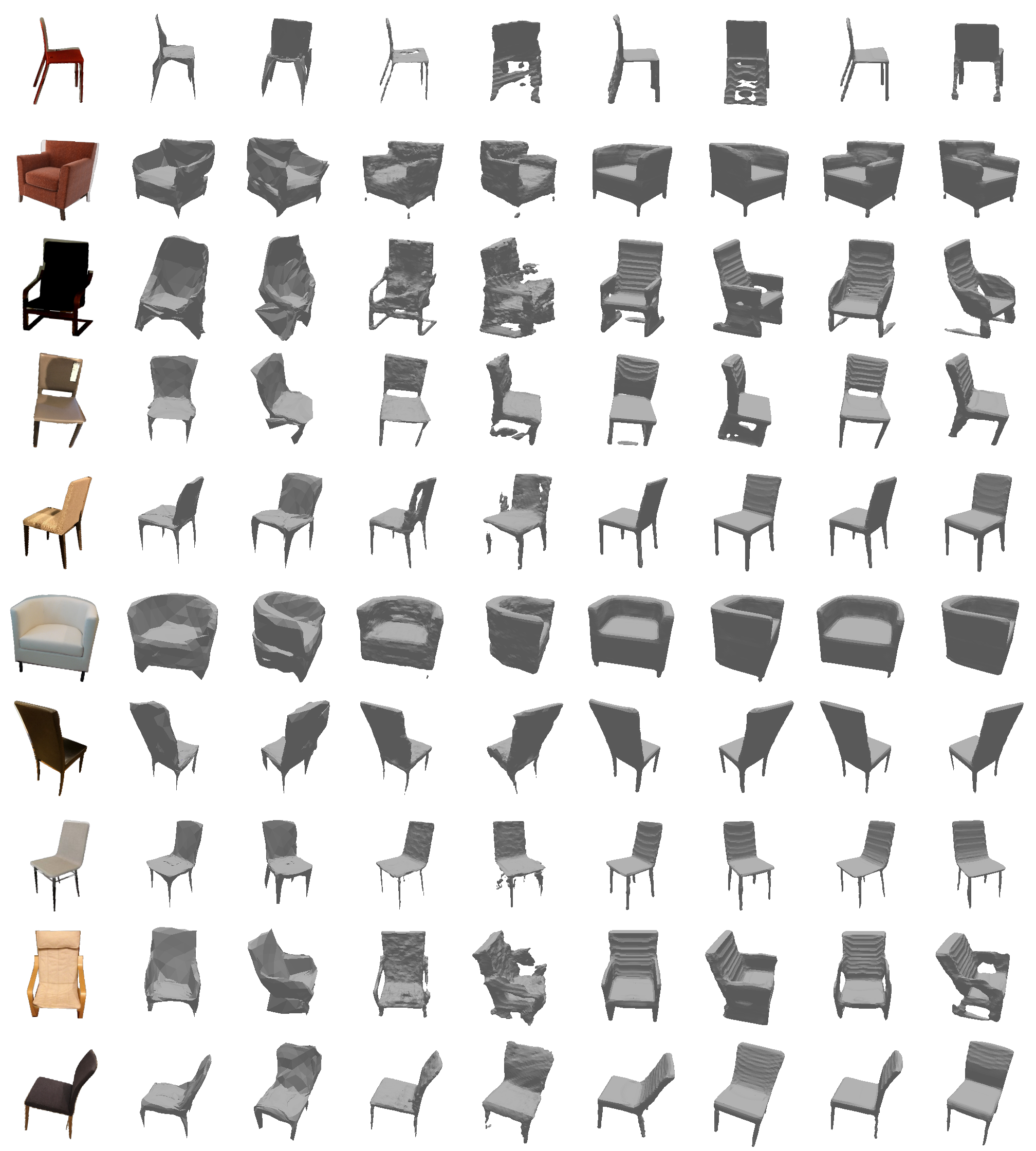}
			\put(2,101){\small{Image }}
			\put(14,101){\small{Pixel2Mesh \cite{Wang_2018}}}
		    \put(36,101){\small{DISN \cite{xu2019disn}}}
			\put(54,101){\small{\textit{MeshSDF}(raw)}}
			\put(75,101){\small{\textit{MeshSDF}}}
			\end{overpic}
		\end{center}
		\vspace{-10pt}
		\caption{\textbf{Comparative results for SVR on Pix3D.} }
\label{fig:Pix3D_qual}
\vspace{-6pt}
\end{figure}
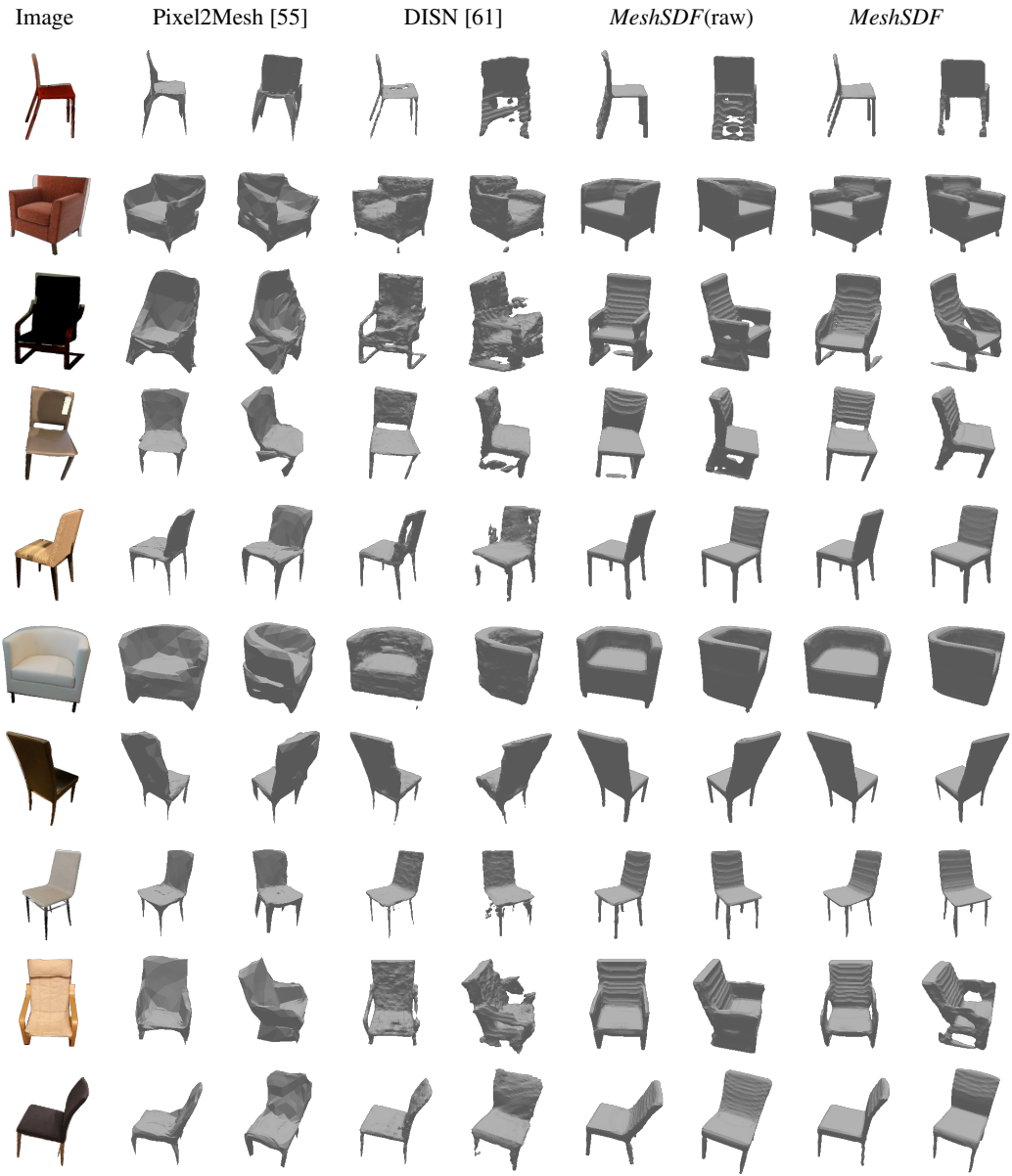
\begin{figure}[t]
        \vspace{20pt}
		\begin{center}
			\begin{overpic}[clip, trim=0.0cm 0cm 0 0.0cm,width=1.0\textwidth]{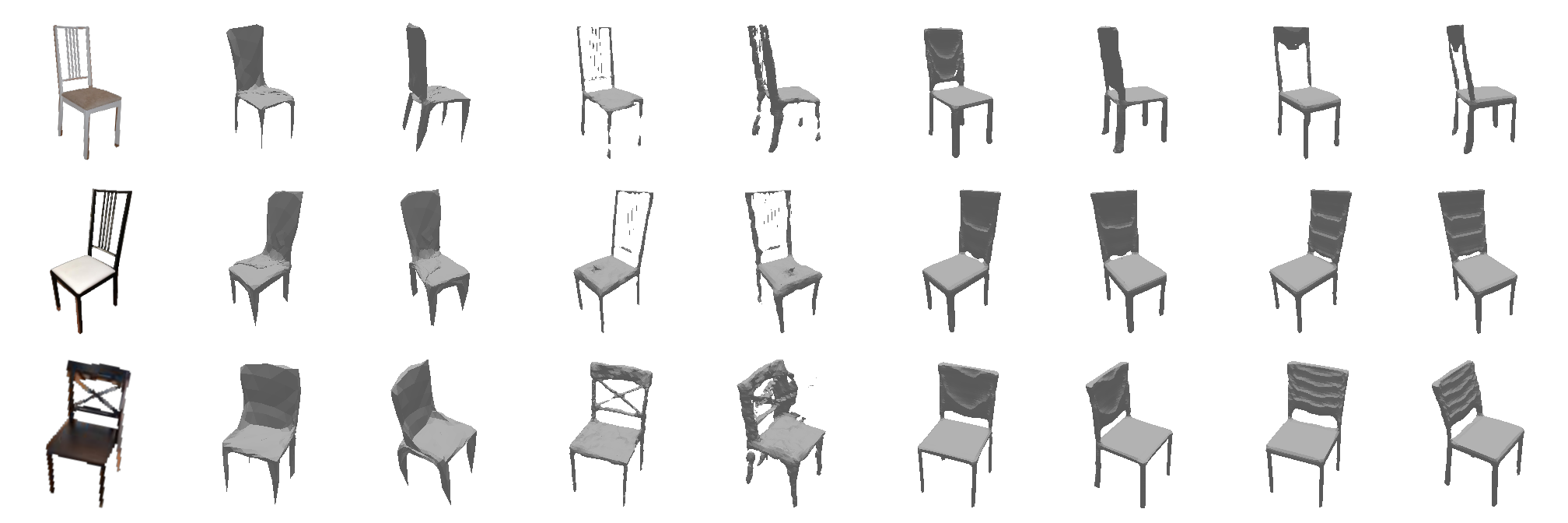}
			\put(2,35){\small{Image }}
			\put(15,35){\small{Pixel2Mesh \cite{Wang_2018}}}
		    \put(39,35){\small{DISN \cite{xu2019disn}}}
			\put(60,35){\small{\textit{MeshSDF}(raw)}}
			\put(85,35){\small{\textit{MeshSDF}}}
			\end{overpic}
		\end{center}
		\vspace{-10pt}
		\caption{\textbf{Failure cases for SVR on Pix3D.} Reconstruction refinement based on $L_1$ silhouette distance fails at capturing fine topological details for challenging samples. In the future, we plan to perform refinement using image-based loss functions that are more sensitive to topological mistakes \cite{Mosinska18}.   }
\label{fig:failure}
\vspace{-10pt}
\end{figure}
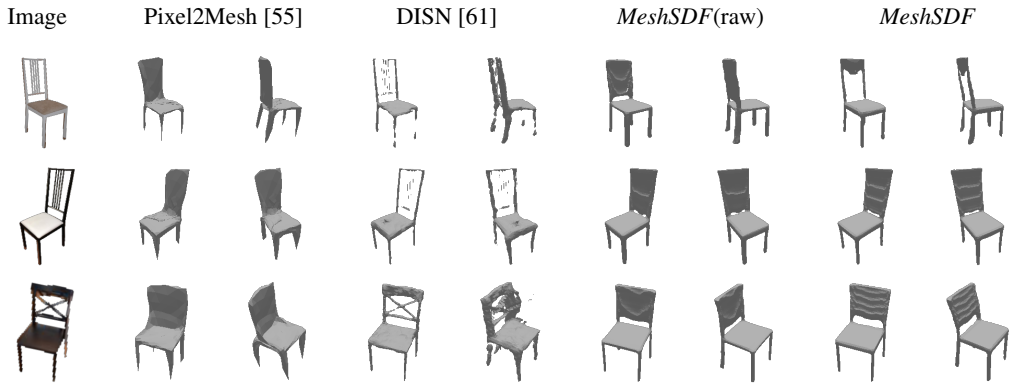
\begin{figure}[h!]
		\begin{center}
			\vspace{-5pt}
			\begin{overpic}[clip, trim=0.0cm 5.0cm 0.0cm 6.0cm,width= 1\textwidth]{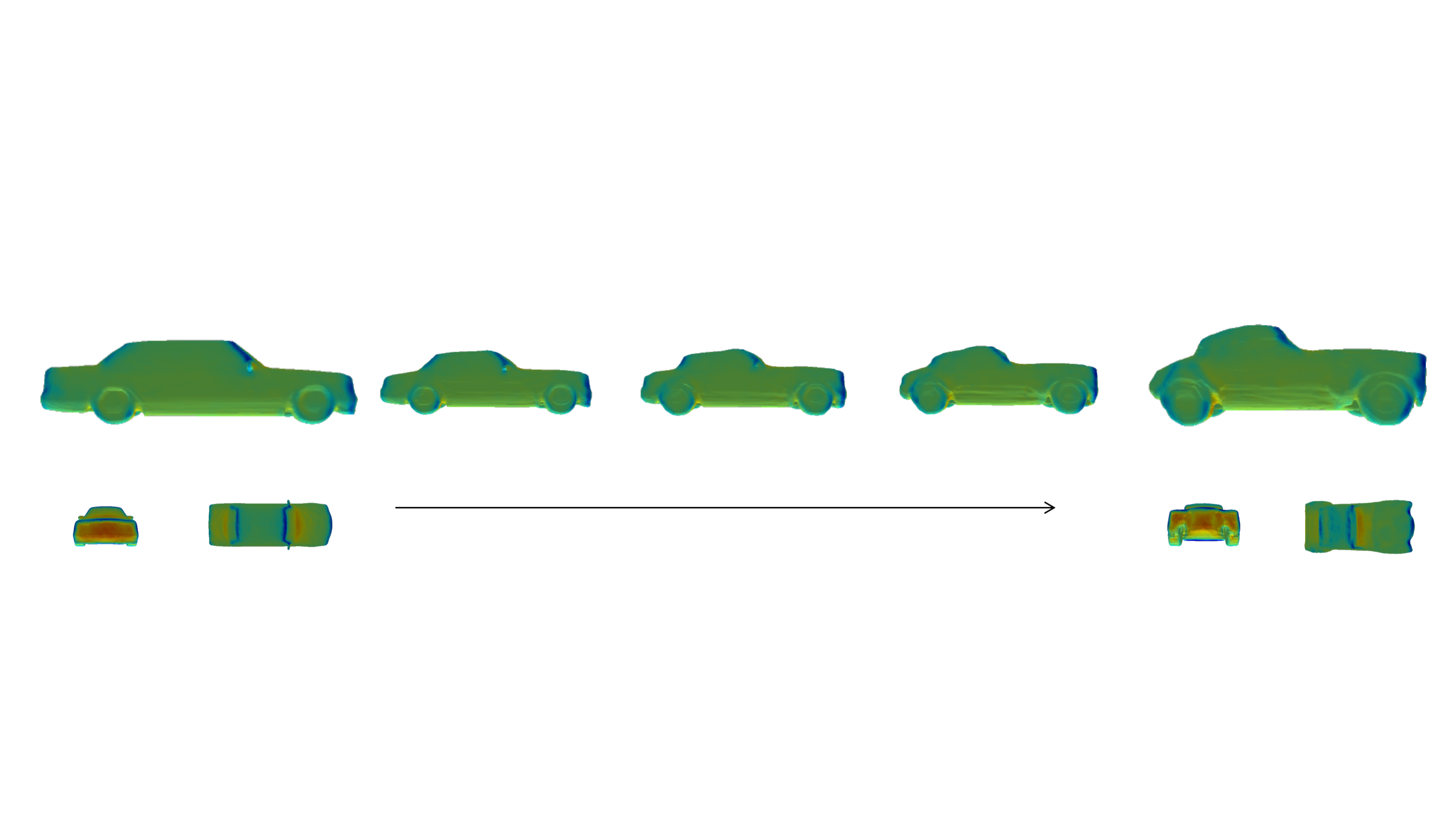}
			\end{overpic}
			\begin{overpic}[clip, trim=0.0cm 6.0cm 0.0cm 5.0cm,width= 1\textwidth]{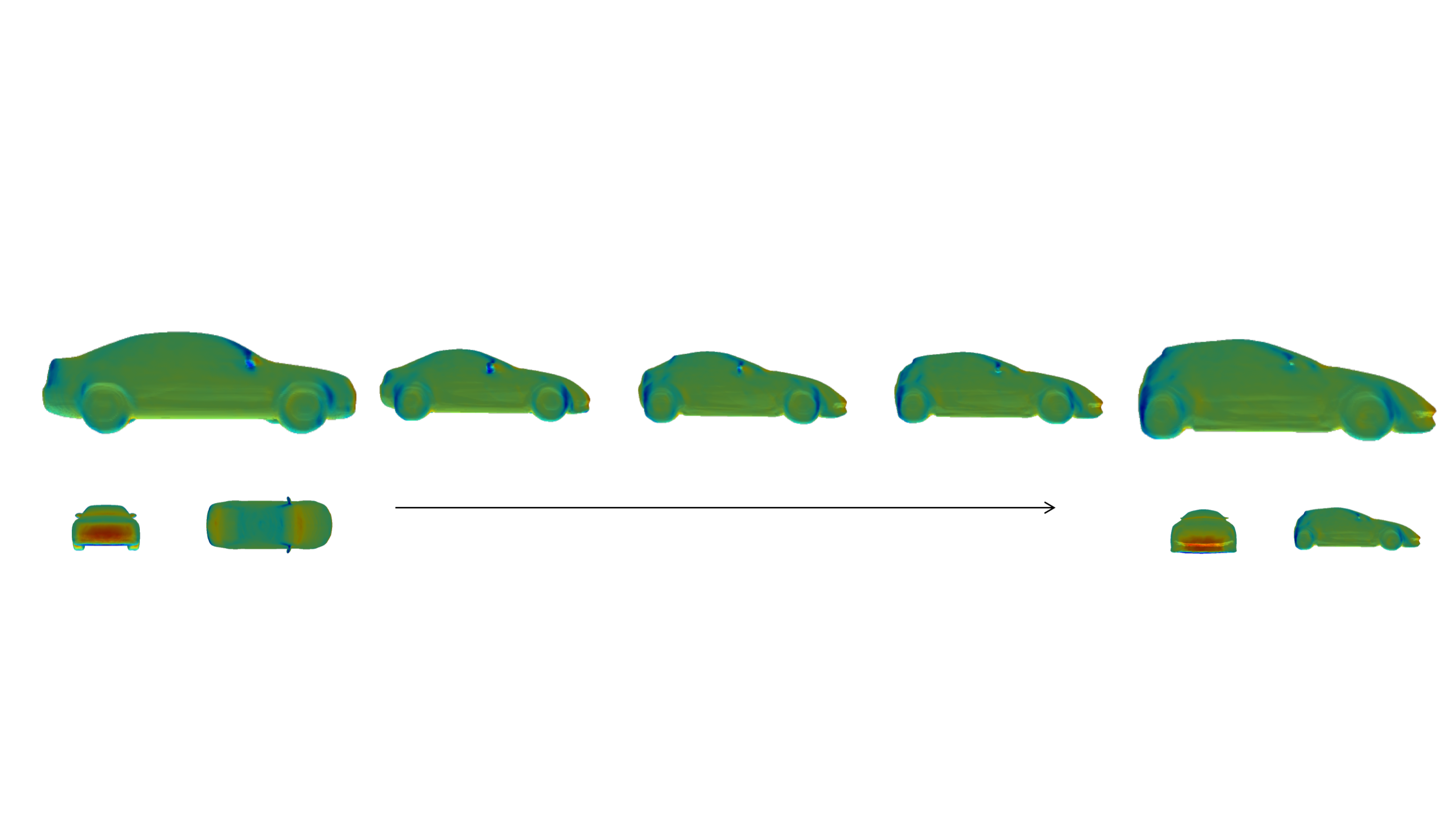}
			\end{overpic}
			\begin{overpic}[clip, trim=0.0cm 6.0cm 0.0cm 5.0cm,width= 1\textwidth]{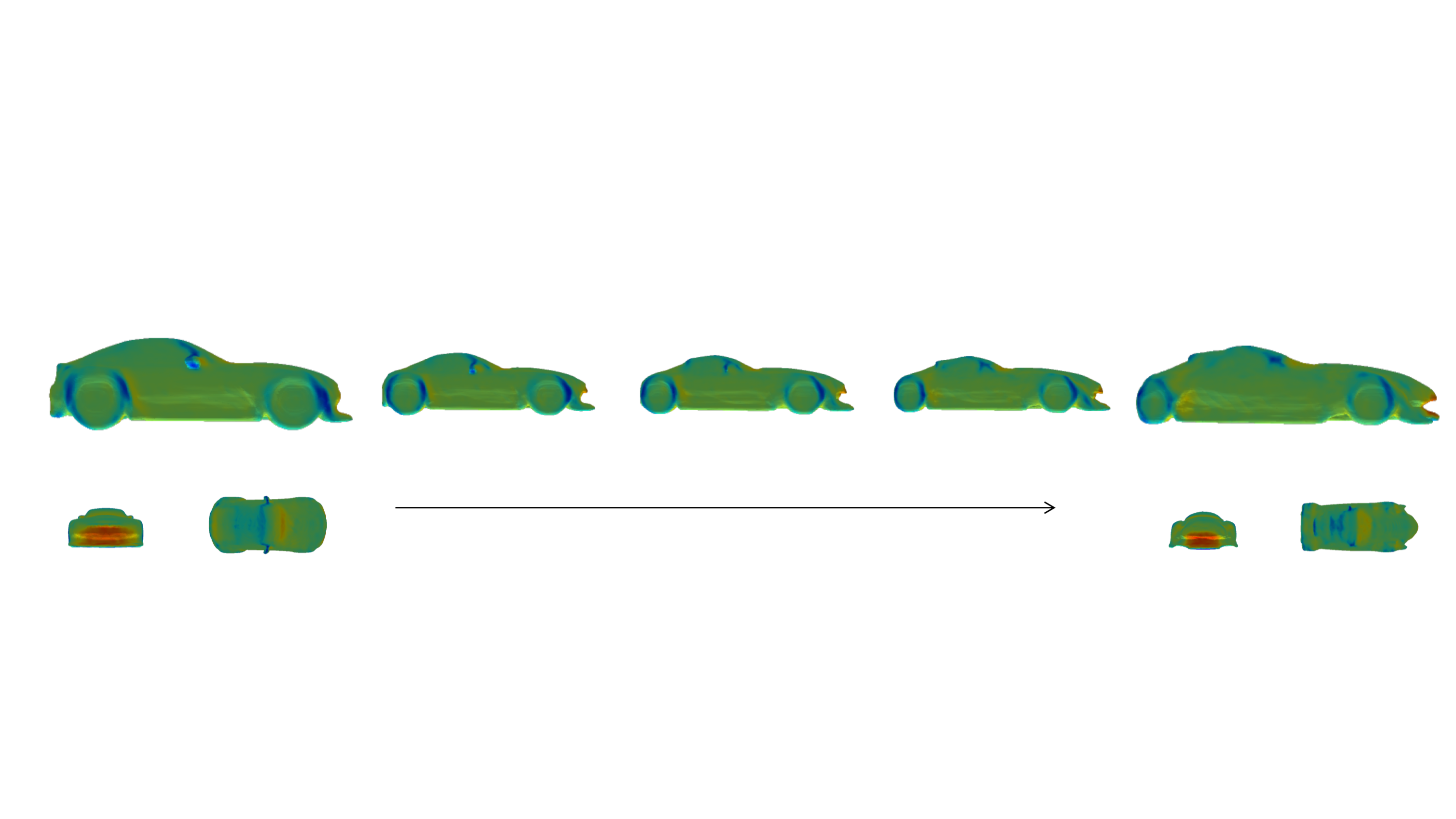}
			\end{overpic}
			\begin{overpic}[clip, trim=0.0cm 6.0cm 0.0cm 5.0cm,width= 1\textwidth]{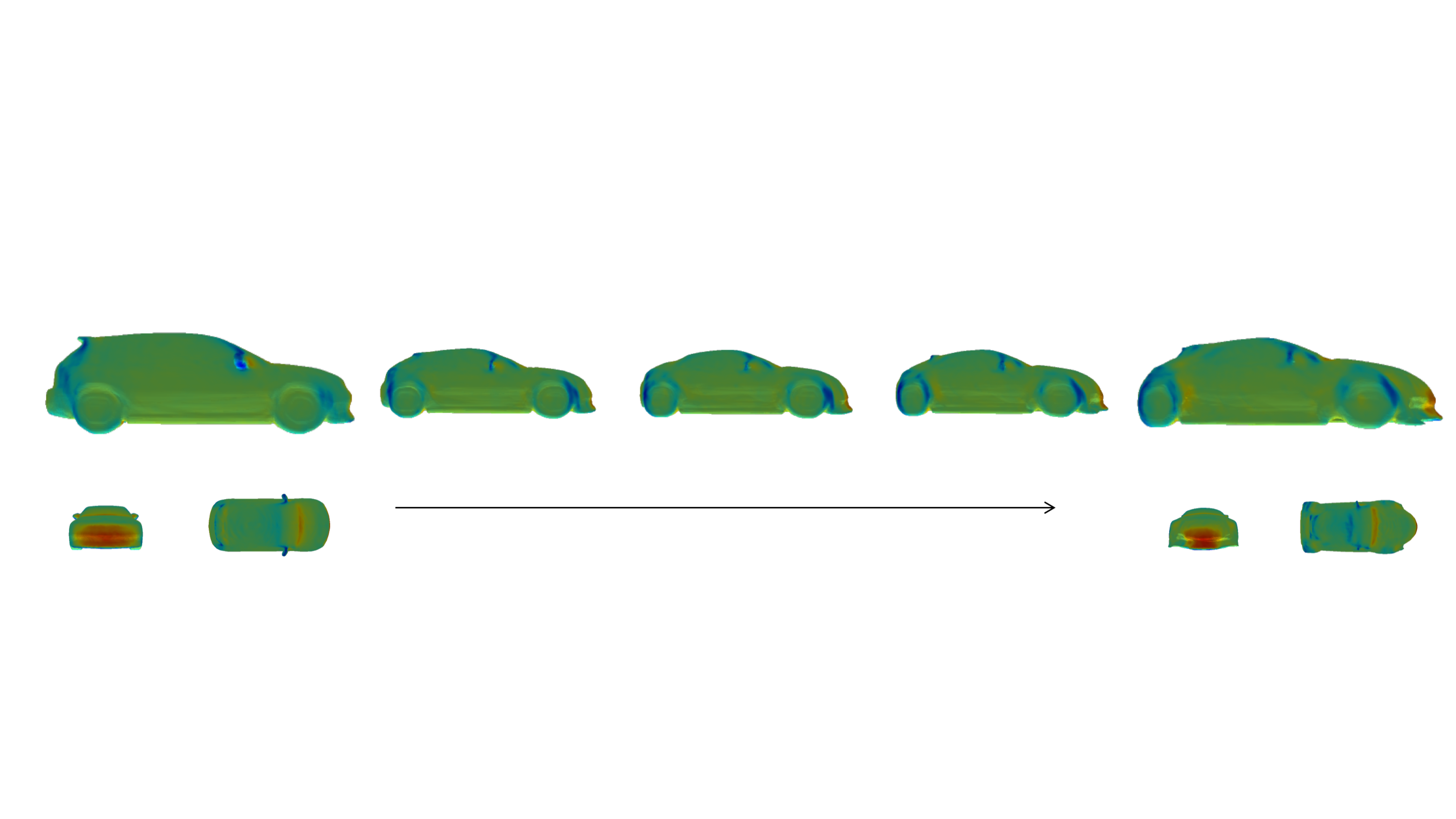}
			\end{overpic}
			\begin{overpic}[clip, trim=0.0cm 6.0cm 0.0cm 5.0cm,width= 1\textwidth]{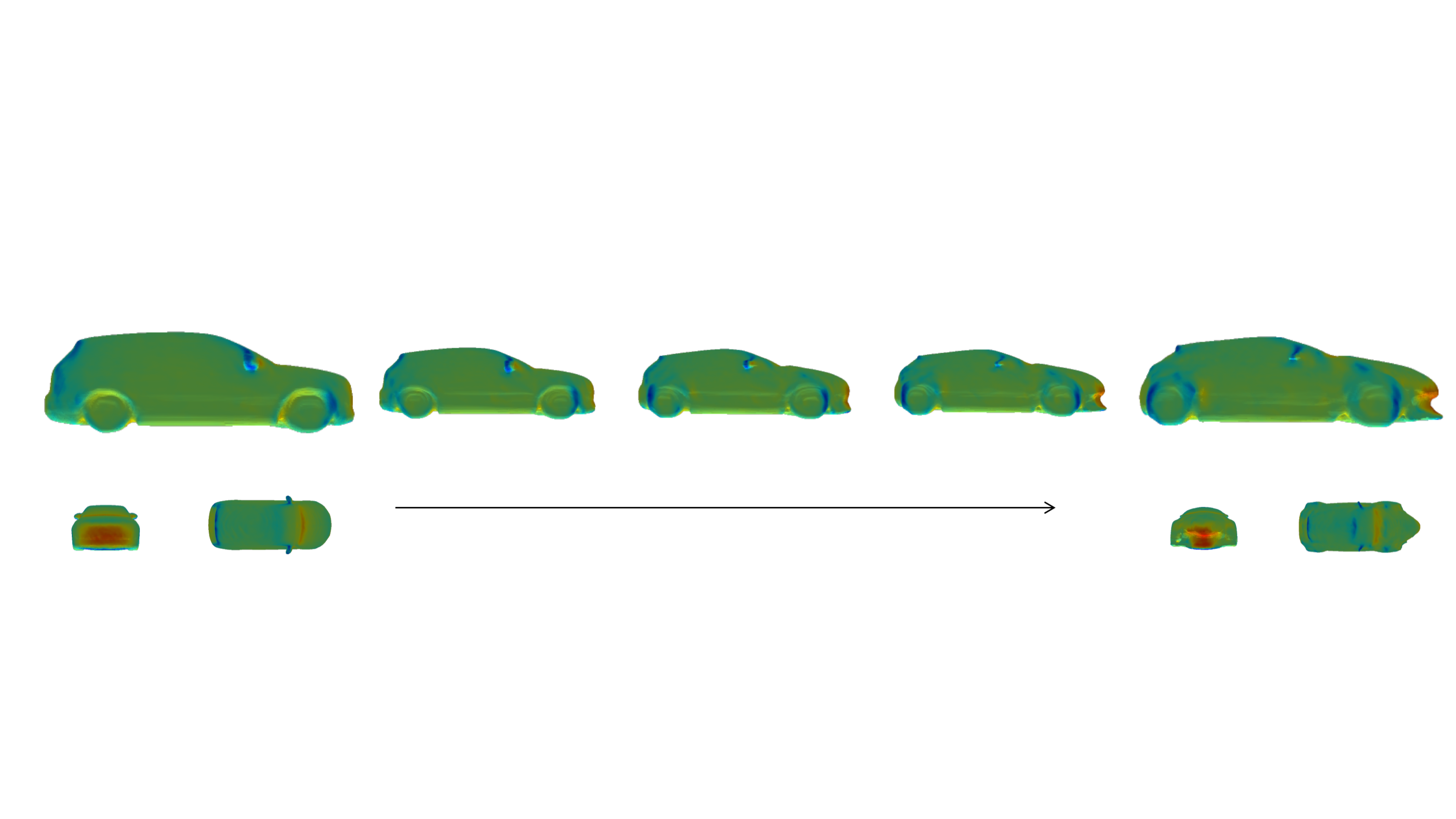}
			\end{overpic}
			\begin{overpic}[clip, trim=0.0cm 6.0cm 0.0cm 5.0cm,width= 1\textwidth]{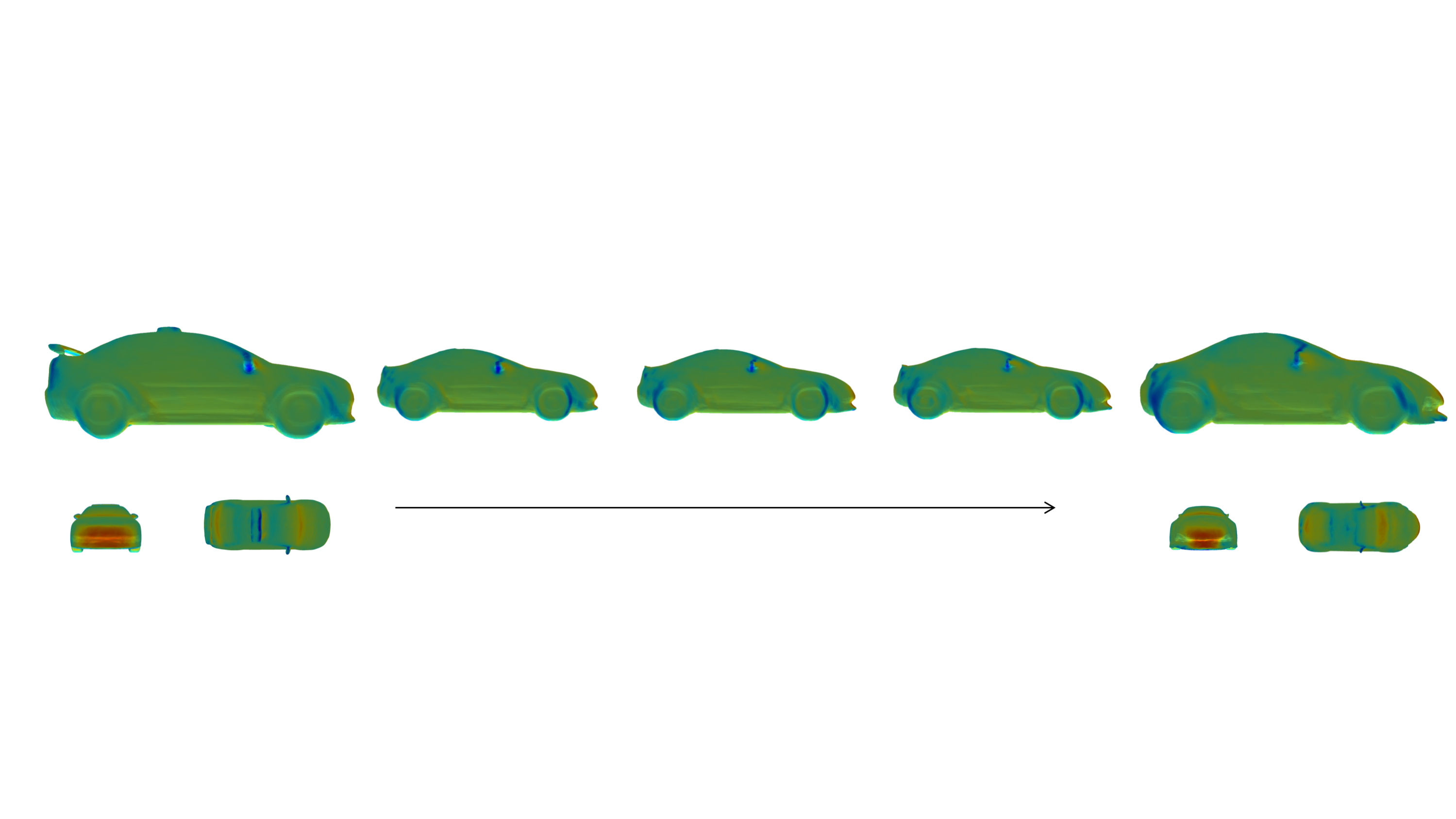}
			\end{overpic}
		\end{center}
		\caption{\textbf{MeshSDF aerodynamic optimizations.}}
\label{fig:cfd_opta}
\end{figure}

\end{document}